\documentclass[pmlr,twocolumn,10pt]{jmlr} %

\usepackage{booktabs}
\usepackage{siunitx}
\usepackage{placeins, bm, bbm}
\usepackage{enumitem, caption}

 \newcommand{\ind}{\perp\!\!\!\!\perp}

\newcommand{\E}{\mathbb{E}}

\theorembodyfont{\upshape}
\theoremheaderfont{\scshape}
\theorempostheader{:}
\theoremsep{\newline}

\DeclareMathOperator*{\argmin}{arg\,min}

\jmlrvolume{LEAVE UNSET}
\jmlryear{2022}
\jmlrsubmitted{LEAVE UNSET}
\jmlrpublished{LEAVE UNSET}
\jmlrworkshop{Conference on Health, Inference, and Learning (CHIL) 2022} %

\title{Improving the Fairness of Chest X-ray Classifiers}

\author{%
\Name{Haoran Zhang} \Email{haoranz@mit.edu}\\
\addr Massachusetts Institute of Technology
\AND
\Name{Natalie Dullerud} \Email{natalie.dullerud@mail.utoronto.edu}\\
\addr University of Toronto
\AND
\Name{Karsten Roth} \Email{karsten.roth@uni-tuebingen.de}\\
\addr University of Tübingen
\AND
\Name{Lauren Oakden-Rayner} \Email{lauren.oakden-rayner@adelaide.edu.au}\\
\addr University of Adelaide
\AND
\Name{Stephen Pfohl} \Email{spfohl@stanford.edu}\\
\addr Stanford University
\AND
\Name{Marzyeh Ghassemi} \Email{mghassem@mit.edu}\\
\addr Massachusetts Institute of Technology
}

\begin{document}

\maketitle

\begin{abstract}
Deep learning models have reached or surpassed human-level performance in the field of medical imaging, especially in disease diagnosis using chest x-rays. However, prior work has found that such classifiers can exhibit biases in the form of gaps in predictive performance across protected groups. In this paper, we question whether striving to achieve zero disparities in predictive performance (i.e. group fairness) is the appropriate fairness definition in the clinical setting, over minimax fairness, which focuses on maximizing the performance of the worst-case group. We benchmark the performance of nine methods in improving classifier fairness across these two definitions. We find, consistent with prior work on non-clinical data, that methods which strive to achieve better worst-group performance do not outperform simple data balancing. We also find that methods which achieve group fairness do so by worsening performance for all groups. In light of these results, we discuss the utility of fairness definitions in the clinical setting, advocating for an investigation of the bias-inducing mechanisms in the underlying data generating process whenever possible. 
\end{abstract}

\paragraph*{Data and Code Availability}
We make use of two chest x-ray datasets: MIMIC-CXR \citep{johnson2019mimic} and CheXpert \citep{irvin2019chexpert}. Both datasets are publicly available pending appropriate data usage agreements. Demographic data for patients in MIMIC-CXR were obtained from MIMIC-IV \citep{mimic_iv}, available through PhysioNet \citep{goldberger2000physiobank}. We analyze an additional radiologist-labelled dataset in this paper. We recruit a board-certified radiologist co-author to manually label 1,200 reports in MIMIC-CXR which have been labelled as \textit{No Finding} by the CheXpert labeller, an automatic rule-based NLP model \citep{irvin2019chexpert}. This dataset, along with code to reproduce our results, can be found at \url{https://github.com/MLforHealth/CXR_Fairness}.

\section{Introduction}
\label{sec:intro}
As machine learning classifiers are becoming increasingly more common in the clinical setting \citep{sendak2020path}, it is important to assess potential potential model biases across protected groups \citep{chen2020ethical}, and, where possible, take measures that minimize the impact of these biases on patient care \citep{vayena2018machine, wiens2019no}. 

In the field of medical imaging, deep learning models have been shown to achieve or even surpass human level performance \citep{liu2019comparison}, e.g. in screening for breast cancer from mammography \citep{mckinney2020international}, macular degeneration from retinal images \citep{burlina2018utility} or pneumonia from chest x-rays \citep{rajpurkar2017chexnet}. However, prior work has found that chest x-ray diagnostic classifiers exhibit significant disparities in the true positive rate (TPR) and false positive rate (FPR) across a variety of datasets, tasks, and protected attributes \citep{seyyed-kalantariCheXclusionFairnessGaps2020}. For example, chest x-ray classifiers trained to detect the presence of any disease significantly underdiagnose Black females, potentially resulting in delays in treatment \citep{seyyed2021medical}.

In order to make fair machine learning models applicable in realistic clinical settings, we must first understand what it means for a classifier to be ``biased'' or ``fair'' in the clinical setting. Most prior work on fairness in healthcare have focused on group fairness \citep{zhang2020hurtful, pfohl2019creating, chen2019can}, which strives to achieve equal performance metrics (e.g. TPR, FPR) between protected groups. However, it is unclear under what conditions such a fairness definition would be appropriate in the clinical setting, over the myriad of other fairness definitions in the machine learning literature, such as minimax fairness \citep{dianaMinimaxGroupFairness2021}, subgroup fairness \citep{kearns2018preventing}, counterfactual fairness \citep{kusner2017counterfactual}, or individual fairness \citep{dwork2012fairness}.

Next, in the case that a machine learning classifier is deemed unfair with respect to some fairness definition, how effective are current computational methods at ``debiasing'' these classifiers in the clinical setting, and how does debiasing with respect to a particular fairness definition affect the disparity in other fairness definition(s)?

In this work, we aim to address these questions on the task of disease classification using chest x-ray images, focusing on group fairness and minimax fairness. We make the following contributions:
\begin{itemize}
    \item We expand upon prior work \citep{seyyed2021medical} to show that Empirical Risk Minimization (ERM, \cite{vapnik1992principles}) models trained to predict a variety of pathologies yield statistically significant performance gaps across many metrics and protected attributes. 
    \item We benchmark a variety of existing methods which aim to improve worst-group performance, and find that no method outperforms simple data balancing.
    
    \item We show, consistent with prior work on tabular data \citep{pfohlEmpiricalCharacterizationFair2021, lahotiFairnessDemographicsAdversarially2020}, that current debiasing methods for group fairness applied to medical images tend to achieve performance parity by worsening performance for all groups.
    \item We provide a preliminary investigation of the possible mechanisms by which disparities in performance metrics can arise in \textit{No Finding} prediction in MIMIC-CXR \citep{johnson2019mimic}, and stress the importance of probing the origin of such disparities in the clinical setting.
\end{itemize}

Finally, in light of our findings, we discuss the utility of the minimax definition of fairness \citep{dianaMinimaxGroupFairness2021} in the clinical setting compared to traditional group fairness definitions that assess differences in error rates across protected groups, and provide recommendations for fair machine learning in the clinical setting.

\section{Background}
\subsection{Fairness Definitions}
Traditional group fairness definitions are specified as conditional independence statements which, in the binary classification setting, entail equality in some performance metric between groups given some threshold \citep{verma2018fairness, hardt2016equality}.  
For example, given the protected group $G$, the binary label $Y$, image $X$, and the binarized prediction $\hat{Y}=\mathbbm{1}[S\geq \tau]$ that results from comparing to a threshold $\tau$ the output $S=h(X)$ of a classifier $h$, equality of odds requires $\hat{Y} \ind G \mid Y$, entailing equal TPR and FPR between groups in the binary classification setting.  
We present several other commonly used definitions of group fairness in Appendix \ref{apd:group_fair}. 

Many group fairness definitions are incompatible with each other (known as ``impossibility theorems'', \cite{barocas-hardt-narayanan, chouldechova2017fair}). Most notably, given an imperfect classifier that outputs a risk score and different base rates between protected groups, it is not possible to have calibration within all groups and equalized odds in the probabilistic sense \citep{kleinbergInherentTradeOffsFair2016, liuImplicitFairnessCriterion2019, pleissFairnessCalibration2017}.

Another definition of fairness which has recently gained popularity is minimax Pareto fairness \citep{ martinezMinimaxParetoFairness2020}. Here, we focus on the pure minimax definition, which is similar to Rawlsian Max-Min fairness \citep{lahotiFairnessDemographicsAdversarially2020}. A classifier $h^*$ over some hypothesis space $\mathcal{H}$ satisfies minimax fairness for some error function $\epsilon$ evaluated on groups $g \in G$ if: \citep{dianaMinimaxGroupFairness2021}
\begin{equation*}
    h^* = \argmin_{h \in \mathcal{H}} \max_{g \in G} \epsilon_g(h)
\end{equation*}

As it is difficult to determine the value of the minima for worst-group performance in practical scenarios, we instead use this as a \textit{relative} definition of fairness. In practice, we say that a classifier $h$ is \textit{fairer} than some baseline classifier $\tilde{h}$ if $\max_{g \in G} \epsilon_g(h) < \max_{g \in G} \epsilon_g(\tilde{h})$.

\subsection{Debiasing Methods} 
There have been a wide array of computational methods developed to debias machine learning models in the binary classification setting. Here, we focus on methods that debias \textit{during training}, and leave methods which debias during preprocessing \citep{louizosVariationalFairAutoencoder2017, wangBalancedDatasetsAre2019, songLearningControllableFair2020} and postprocessing \citep{pleissFairnessCalibration2017, hardt2016equality, kimMultiaccuracyBlackBoxPostProcessing2018} as future work. 

Debiasing methods can strive to achieve group fairness in several ways. First, this can be done by enforcing the appropriate conditional independence with the use of an adversary \citep{edwards2015censoring, wadsworthAchievingFairnessAdversarial2018, zhangMitigatingUnwantedBiases2018, madrasLearningAdversariallyFair2018}. Second, a term can be added to the loss function which corresponds to the distance between distributions to be equalized (e.g. between a group and the marginal) \citep{pfohlEmpiricalCharacterizationFair2021}. Finally, one could also solve the constrained optimization problem using the Lagrangian \citep{cotterOptimizationNonDifferentiableConstraints, lokhandeFairALMAugmentedLagrangian2020}. 

Alternatively, one can instead aim to improve the performance of the worst-case group. GroupDRO \citep{sagawaDistributionallyRobustNeural2020} attempts to minimize the training loss of the worst-case group by exponentially upweighting groups with higher loss after each step. Methods to improve worst-case loss may also be group-unaware. Such methods typically seek to minimize worst-case error over all possible subgroups of a certain size  \citep{duchiDistributionallyRobustLosses, martinezBlindParetoFairness2021}, or instead upweight poorly performing samples during training \citep{namLearningFailureTraining2020, liuJustTrainTwice2021, lahotiFairnessDemographicsAdversarially2020}. For example, Just Train Twice \citep{liuJustTrainTwice2021} first learns a model through empirical risk minimization as usual, and then learns a second model which upweights samples misclassified by the first model. 

In this work, we benchmark the performance of algorithms which seek to achieve group fairness or maximize performance of the worst-case group, by selecting representative methods from each computational approach.

\subsection{Fairness in Computational Medical Imaging}

As machine learning models become increasingly integrated in the healthcare setting, one primary concern is whether such models are being used in a fair and ethical way \citep{ahmad2020fairness, wawira2021equity, chen2020ethical}. In the field of machine learning for medical imaging, there have been several prior works that benchmark the degree of disparities between protected groups for machine learning models. \cite{seyyed-kalantariCheXclusionFairnessGaps2020} demonstrates disparities in TPR between protected groups defined by sex, race, insurance, and age for three publicly available chest x-ray datasets on models trained to predict disease status. \cite{seyyed-kalantariMedicalImagingAlgorithms2021} focuses on the task of \textit{No Finding} prediction, concluding that significant disparities exist in FPR between protected groups (corresponding to underdiagnosis), with the bias often disfavoring historically disadvantaged groups. \cite{larrazabal2020gender} found in the disease classification using chest x-ray setting that decreasing the number of samples in the training set for a protected group often leads to reduced performance for the group. \cite{banerjee2021reading} found that chest x-rays inherently contain racial information, though it is unclear what implications this has for downstream classifier fairness. 

Similar benchmarking studies have been done on dermatology datasets \citep{kinyanjui2020fairness} and CT scans \citep{zhou2021radfusion}, though to our knowledge, our work is the first to benchmark algorithms for bias reduction in the medical imaging setting.

\section{Methods}
We benchmark the performance of the following simple baseline methods:
\begin{itemize}
    \item Empirical Risk Minimization (\texttt{ERM}, \cite{vapnik1992principles}) minimizes population risk irrespective of group compositions.
    \item \texttt{Balanced ERM} upsamples minority groups to minimize risk in a population where groups are equal in size.
    \item \texttt{Stratified ERM} learns a separate model for each protected group.
\end{itemize}

We benchmark the performance of the following methods which try to achieve group fairness. Here, we focus on equalized odds, as it has been used in prior work studying fairness in healthcare \citep{zhang2020hurtful}. We upsample minority groups to ensure equal presence of each protected group in each minibatch. We note that the methods described below do not explicitly seek to improve the performance of any group.

\begin{itemize}
    \item \texttt{Adversarial} \citep{wadsworthAchievingFairnessAdversarial2018} uses an adversary to enforce $S \ind G \mid Y$. 
    \item \texttt{MMDMatch} \citep{pfohlEmpiricalCharacterizationFair2021} penalizes the Maximum Mean Discrepancy (MMD, \cite{gretton2012kernel}) distance between $P(S \mid Y = y, G = g)$ and $P(S \mid Y = y)\ \forall \ g\in G $ and $y \in \{0, 1\}$.
    \item \texttt{MeanMatch} \citep{pfohlEmpiricalCharacterizationFair2021} penalizes the mean of the distributions $P(S \mid Y = y, G = g)$ and $P(S \mid Y = y)\ \forall \ g\in G $ and $y \in \{0, 1\}$.
    \item \texttt{FairALM} \citep{lokhandeFairALMAugmentedLagrangian2020} uses an augmented Lagrangian method to enforce fairness constraints.
\end{itemize}

Finally, we benchmark the following methods which seek to improve the performance of the worst-case group:
\begin{itemize}
    \item \texttt{GroupDRO} \citep{sagawaDistributionallyRobustNeural2020} exponentially upweights groups with worse loss after each minibatch. 
    \item \texttt{ARL} \citep{lahotiFairnessDemographicsAdversarially2020} is a group-unaware method which weights each sample with an adversary that tries to maximize the weighted loss.
    \item \texttt{JTT} \citep{liuJustTrainTwice2021} is a group-unaware method which trains an additional classifier that upweights samples classified incorrectly by the \texttt{ERM} classifier.
\end{itemize}

\section{Experiments}
\label{sec:experiments}
\paragraph{Data} We use all chest x-ray images from MIMIC-CXR \citep{johnson2019mimic} and CheXpert \citep{irvin2019chexpert}. Further summary statistics about the datasets can be found in Appendix Table \ref{tab:cohort_stats_1}.

\paragraph{Protected Groups} We define protected groups based on the following demographic attributes: (1) self-reported race and ethnicity, (2) sex, (3) age, discretized into four intervals. 

\paragraph{Targets} We train separate binary classification models to predict each of the following targets: (1) \textit{No Finding}, corresponding to absence of any pathology, (2) \textit{Pneumothorax}, (3) \textit{Fracture}. Statistics on the prevalence of each of the targets among protected groups can be found in Appendix Table \ref{tab:cohort_stats_2}. For \textit{Fracture} and \textit{Pneumothorax}, we treat samples labelled with an uncertain label as a negative sample. Note that the uncertain label does not exist for \textit{No Finding}. All labels are derived from free-form radiology notes using the CheXpert labeller, a rule-based NLP model \citep{irvin2019chexpert}.

\paragraph{Training} We use an ImageNet-pretrained \citep{deng2009imagenet} DenseNet-121 \citep{huang2017densely}, but replace the final layer with a 2-layer neural network with 384 hidden units and 1 output unit. We split each dataset into a 16.7\% test set, and an 83.3\% cross-validation set split into 5 folds. We train 5 models for each hyperparameter setting, using each fold for model selection and early stopping with the remaining 4 folds left for training. We select the hyperparameter setting with the best validation worst-group AUROC, averaged across the folds. Additional training details can be found in Appendix \ref{apd:train_det}.

\paragraph{Metrics} We evaluate a variety of metrics on the test set for each protected group. 

\textit{Threshold-free Metrics} Though real-world decision making often requires a threshold to make a binary decision, the actual value of the threshold depends highly on the use-case of the model and the preference of the clinician. Absent of such information, we use the following metrics which do not require an operating threshold:
\begin{itemize}
    \item Area under the ROC curve (\texttt{AUROC}) -- the standard metric used to evaluate performance of chest x-ray classifiers, which has also been used in prior fairness analyses \citep{larrazabal2020gender}.
    \item Binary cross-entropy (\texttt{BCE}) -- the metric that is being directly optimized in the loss function.
    \item Expected calibration error (\texttt{ECE}, \cite{nixon2019measuring}). Equal calibration curves imply group sufficiency, a commonly used group fairness definition. Low calibration error is also important as it ensures that, for a particular choice of threshold on the predicted risk score, that we are actually operating at a similar threshold on the true risk for each group.
\end{itemize}

\textit{Threshold-required Metrics} We evaluate the recall and specificity at a fixed threshold. We note that threshold selection is a non-trivial task that requires considering the real-world cost of misclassified samples, and is thus highly application-specific \citep{kording2007decision, bakalarFairnessGroundApplying2021}. Here, we select nominal values for demonstration purposes, and use the same threshold for all groups. We also note the trade-off between recall and specificity. In the event that one group has larger recall but lower specificity than another, it is unclear which group is disadvantaged without considering misclassification costs.

\textit{Threshold-implicit Metrics} We finally evaluate the recall (TPR) at $k\%$ specificity (TNR), as it is often easier to state a desired false positive rate than a desired operating threshold. However, we note that this metric still only assesses performance at a single point of the risk distribution.

\paragraph{Bootstrapping} For each hyperparameter setting, we construct a 95\% confidence interval for all metrics by bootstrapping over samples on the test set and over the 5 trained models. We also construct bootstrapped confidence intervals for $\epsilon(h) - \epsilon(\tilde{h})$. We choose $\tilde{h}$ as the \texttt{Balanced ERM} classifier, as it has been shown to be competitive on worst-group performance in prior non-clinical work \citep{idrissiSimpleDataBalancing2021}.

\begin{figure*}[]
    \centering
    \hspace*{-1.cm}\includegraphics[width=1.15\textwidth]{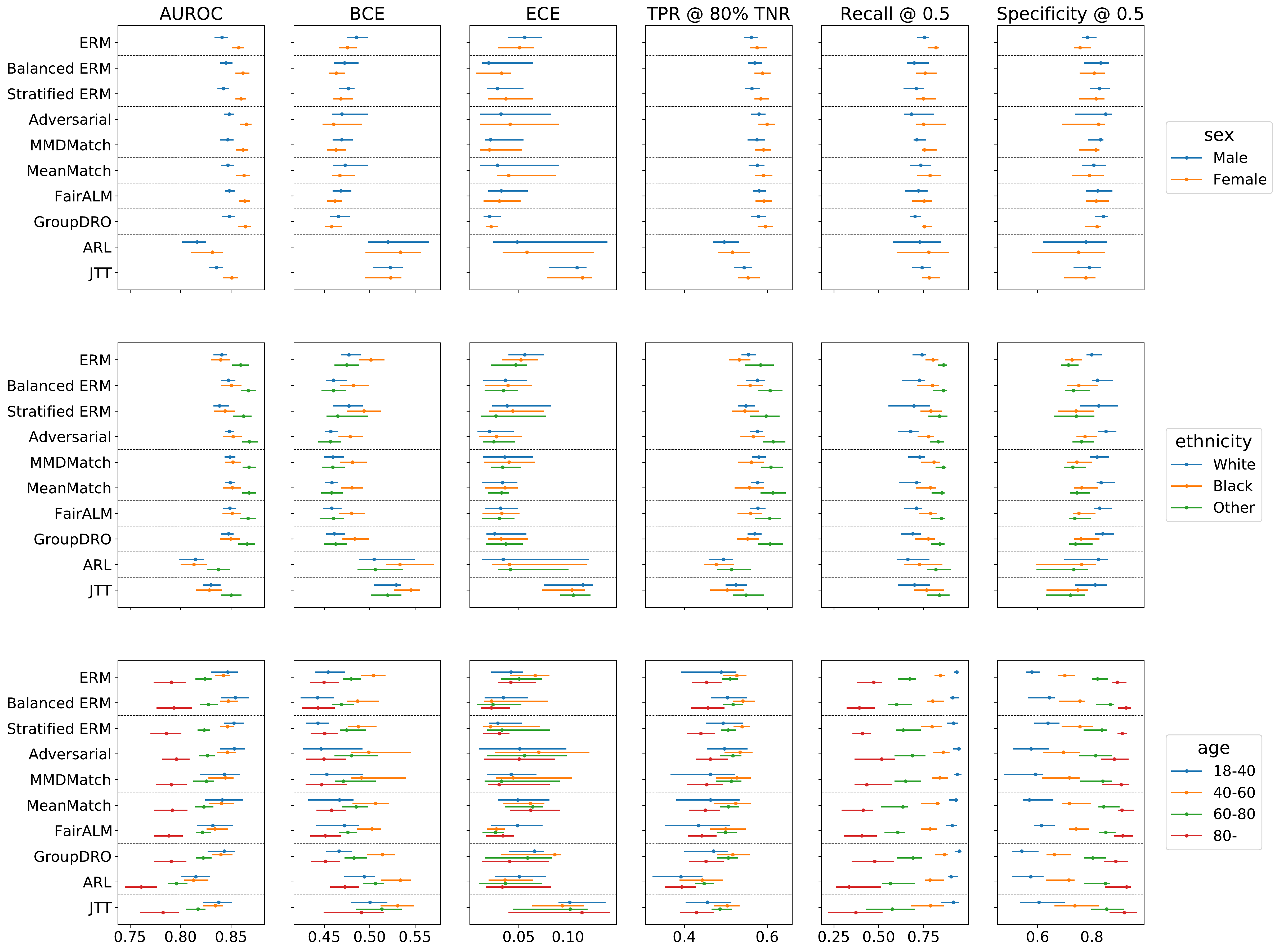}
    \caption{Comparison of the performance of models that predict \textit{No Finding} in MIMIC-CXR. Error bars indicate 95\% confidence intervals from 250 bootstrap iterations. We note that significant performance gaps exist between protected groups, and such gaps can vary depending on the metric examined.}
    \label{fig:mimic_compr}
\end{figure*}

\section{Results}
\label{sec:results}

We report results for \textit{No Finding} prediction in MIMIC-CXR in the main paper, as they are representative of the results obtained from other tasks and datasets, and has been the subject of study in prior fairness work \citep{seyyed-kalantariMedicalImagingAlgorithms2021}. We present similar figures for the remaining tasks and datasets in Appendix \ref{apd:results}.

\subsection{Significant Performance Gaps}

In Figure \ref{fig:mimic_compr}, we compare the performance of models trained to predict \textit{No Finding} in MIMIC-CXR. Focusing on the \texttt{ERM} model, our results show that significant performance gaps exist across many metrics for all of the protected attributes examined. Such gaps are especially large across different age groups -- specifically, all models perform much worse in older populations as seen in the AUROC results. We note that the directionality of the gap observed can vary greatly between metrics. For example, as noted in \cite{seyyed-kalantariMedicalImagingAlgorithms2021} for an \texttt{ERM} model, Black patients have lower specificity (higher underdiagnosis) at a fixed threshold compared to White patients, but larger recall. However, we observe that the \texttt{ERM} model does not exhibit significant differences in AUROC or calibration error between Black and White patients.

\subsection{No Method Outperforms Balanced-ERM on Worst-Case Group}

\begin{figure*}[]
    \centering
    \hspace*{-1cm}\includegraphics[width=1.15\textwidth]{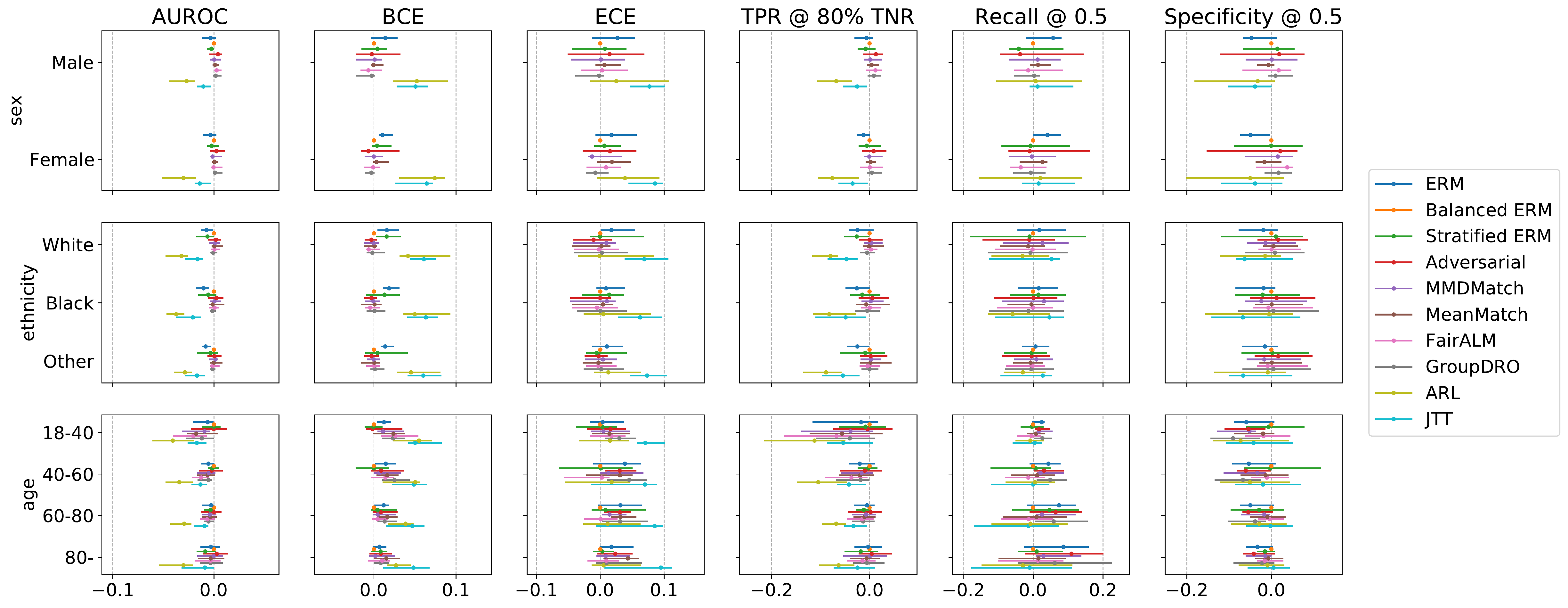} 
    \caption{Comparison of models that predict \textit{No Finding} in MIMIC-CXR. We show the difference in performance between each model and \texttt{Balanced ERM}. Positive values indicate that a model has a higher value for a metric than \texttt{Balanced ERM}.  Error bars indicate 95\% confidence intervals from 250 bootstrap iterations. We observe that no method significantly outperforms \texttt{Balanced ERM}. }
    \label{fig:mimic_delta}
\end{figure*}

In Figure \ref{fig:mimic_delta}, we compare the performance of all models with the performance of \texttt{Balanced ERM}. We find, similar to prior work on non-clinical data \citep{idrissiSimpleDataBalancing2021}, that no model outperforms simple data balancing on any evaluation metric. Specifically, we note that \texttt{JTT} and \texttt{ARL} seem to give worse-performing models that are more poorly calibrated, and so do not achieve the desired goal of improving worst-group performance. Similar to prior work on tabular data \citep{pfohl2021comparison}, we find that \texttt{GroupDRO} also does not significantly improve worst-group performance. However, we do note that there are instances where \texttt{Balanced ERM} outperforms \texttt{ERM} and \texttt{Stratified ERM}, especially in AUROC, which may be connected with prior work showing that increasing number of samples for a protected group increases its performance in the chest x-ray setting \citep{larrazabal2020gender}.

\subsection{Group Fairness Worsens All Groups}

\begin{figure*}[]
    \centering
    \hspace*{1.0cm}\includegraphics[width=1\textwidth]{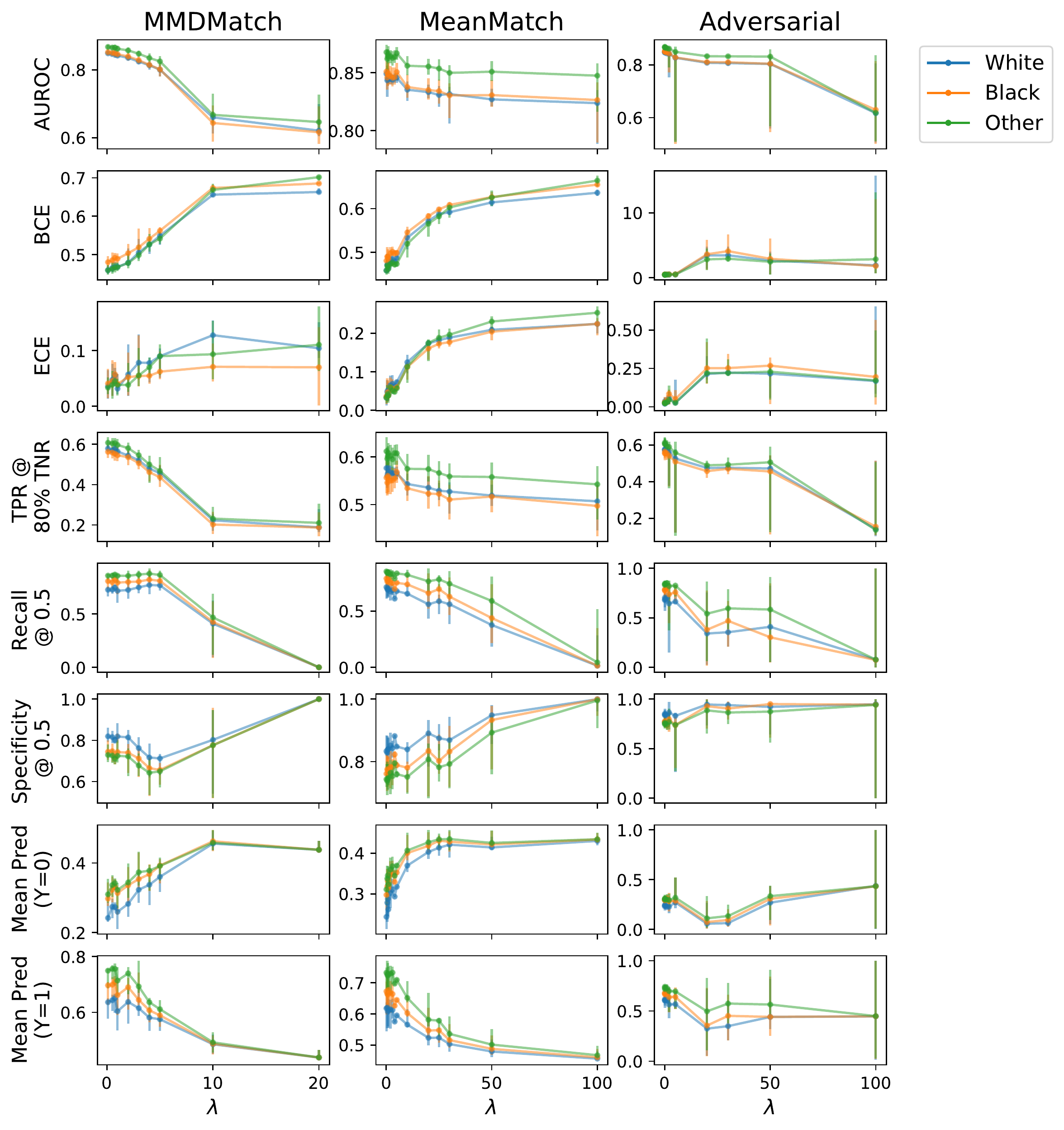} 
    \caption{Comparison of models that predict \textit{No Finding} in MIMIC-CXR which trying to achieve equalized odds between ethnicities, plotted as a function of $\lambda$, the weighting of the additional loss term. Error bars indicate 95\% confidence intervals from 250 bootstrap iterations. We observe that all models achieve equalized odds with large $\lambda$, though no model improves performance metrics for any group, and all models significantly worsen calibration. Note that the range of $\lambda$ varies between methods, as the additional loss terms do not have the same scale.}
    \label{fig:compr_adv_metric}
\end{figure*}

In Figure \ref{fig:mimic_compr}, we observe that the group fairness methods do not seem to be closing the gap in disparities in TPR and FPR when the worst-group AUROC is used as the selection metric. To further examine this behavior, we plot in Figure \ref{fig:compr_adv_metric} a variety of performance metrics for three methods that add an additional term to the loss function (\texttt{Adversarial}, \texttt{MMDMatch}, \texttt{MeanMatch}) in order to achieve equalized odds, as a function of the weighting of the additional loss term during training ($\lambda$). When $\lambda = 0$, all three methods are equivalent to \texttt{Balanced-ERM}. We additionally include the mean prediction for the positive samples (i.e. the mean of the distribution $P(S \mid G=g, Y=1)$) and negative samples. Equality of this metric corresponds to the probabilistic version of equalized odds, and is the quantity directly penalized in \texttt{MeanMatch}. 

We first observe that for all three of the methods, increasing $\lambda$ does not result in an increase in any of the performance metrics for any of the groups, and any increase in recall is met with a corresponding decrease in specificity. This explains the behavior observed in Figure \ref{fig:mimic_compr}, where only model with low penalty are shown due to the model selection criteria. 

For all three methods, we do achieve equalized odds (in both the binary and probabilistic sense) for large values of $\lambda$, as the gap in the recall, specificity, and mean prediction go to zero. However, at those large values of $\lambda$, the model is degraded significantly over the \texttt{ERM} model, as we observe large drops in AUROC and increases in BCE and ECE. 
The worsening of model calibration to achieve equalized odds supports previous work showing the incompatibility between the two definitions \citep{kleinbergInherentTradeOffsFair2016, liuImplicitFairnessCriterion2019}.%

In any case, we observe, similar to prior work on tabular \citep{pfohlEmpiricalCharacterizationFair2021} and non-clinical \citep{lahotiFairnessDemographicsAdversarially2020} datasets, that enforcing group fairness constraints result in reduced model performance for all protected groups.

\section{Dissecting the Source of Bias}
\label{sec:dissecting_bias}
There are many potential dataset biases that could arise during a data collection process \citep{gianfrancesco2018potential}. When a machine learning model is evaluated on biased test-set data, the results obtained may not be reflective of real-world model performance \citep{mehrabi2021survey, yu2020one, wickUnlockingFairnessTradeoff}. In Section \ref{sec:results}, absent of additional information about biases in the data generating process, we assumed that the test set distribution is reflective of the deployment setting distribution. Here, we examine whether this assumption is valid for the \textit{No Finding} task in MIMIC-CXR. We focus on \textit{label bias} in this work, and leave other potential sources of bias for future work. %

\subsection{Radiologist-Labelled Samples}

\paragraph{Motivation} Labels in MIMIC-CXR are extracted from free-text radiology reports using the CheXpert labeller \citep{irvin2019chexpert}, a rule-based and expertly-defined NLP model. Evaluations of the CheXpert labeller for the \textit{No Finding} task against gold standard radiologist labels in prior work have uncovered surprisingly poor label accuracy: 54.3\% F1 score on MIMIC-CXR \citep{smit2020chexbert} and 76.9\% F1 score on CheXpert \citep{irvin2019chexpert}. Here, we first validate the poor performance of the automatic labeller using a new set of radiologist labelled samples. Then, we examine whether the degree of label bias differs between intersectional subgroups, and correlate such disparities with the gaps seen in Figure \ref{fig:mimic_compr}.

\paragraph{Setup} We select 1,200 radiology reports in MIMIC-CXR that have been labelled as \textit{No Finding} by the automatic labeller, corresponding to roughly 200 samples each from the intersections of sex and ethnicity. We recruit a board-certified radiologist co-author to verify whether each report actually indicates \textit{No Finding} using only the free-form text, without access to the underlying chest x-ray or any other patient information. 

\begin{figure}[]
    \centering
    \includegraphics[width=0.95\linewidth]{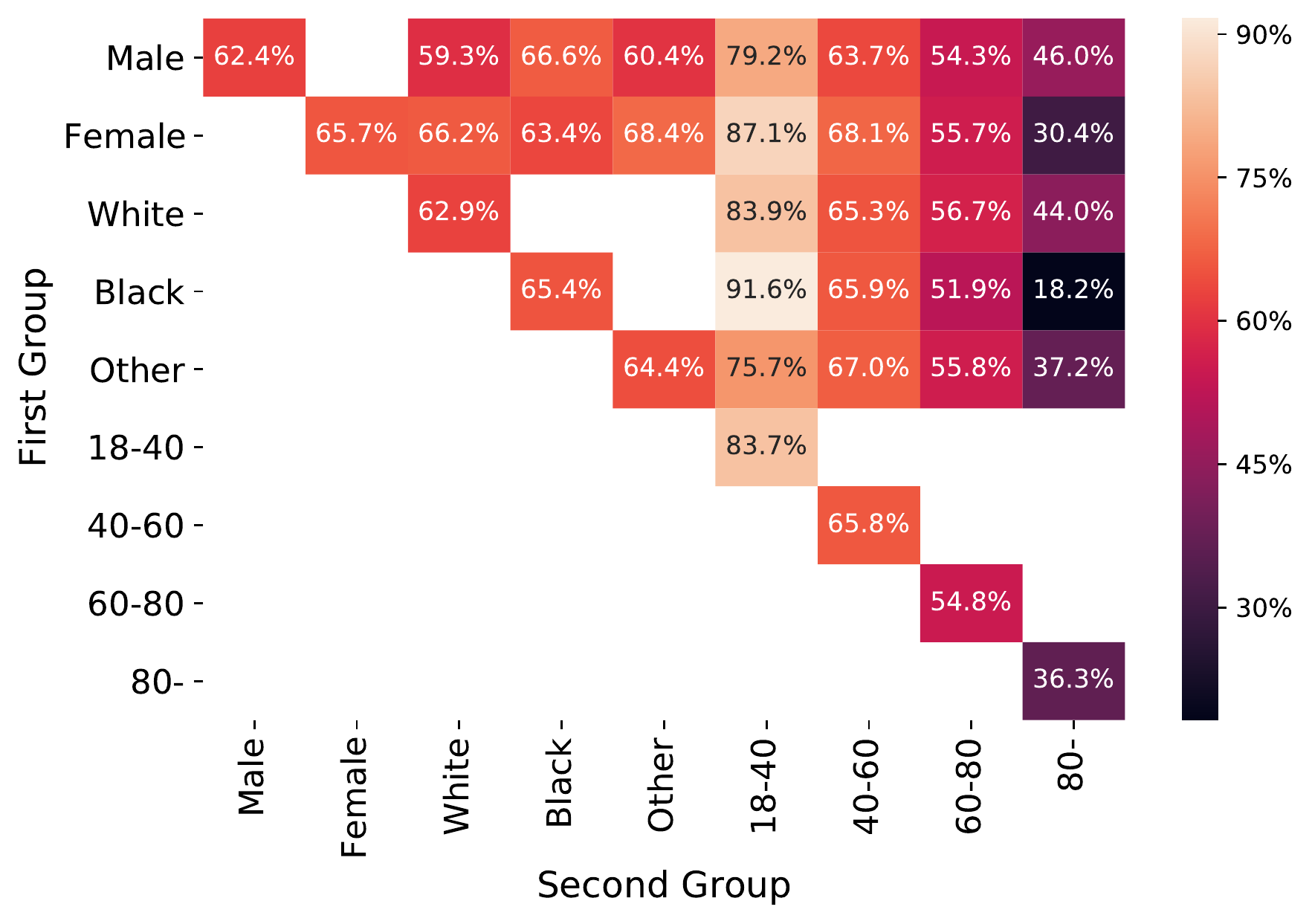} 
    \caption{Accuracy of the CheXpert labeller on 1,200 radiology reports in MIMIC-CXR which it labels as \textit{No Finding} relative to the radiologist gold standard, for each protected group and intersectional subgroup. }
    \label{fig:rad_labels_without_CIs}
\end{figure}

\paragraph{Results} In Figure \ref{fig:rad_labels_without_CIs}, we report the accuracy of the CheXpert labeller for each protected group and intersectional subgroup, assuming that the radiologist labels are the gold standard. Each cell in the heatmap corresponds to the probability that a group has \textit{No Finding}, given that the CheXpert labeller labels it as so (i.e. a positive predictive value). A version of this figure with 95\% confidence intervals can be found in Appendix Figure \ref{fig:rad_labels_with_CIs}. 

We find that the quality of the CheXpert labeller is poor across the board. Overall, when the labeller labels a report as \textit{No Finding}, it is only correct 64.1\% of the time. Looking at the accuracy for each group, we find no significant differences in the label quality between sexes and ethnicities, or their intersections. However, there are significant disparities between age groups -- specifically, those in the "80-" group have the worst-quality labels, and those in the "18-40" group have the best. Interestingly, this also correlates with the performance disparities observed in Figure \ref{fig:mimic_compr}, indicating that label bias may be responsible for poor performance of models in the "80-" group. We note that these results may be affected by age-related comorbidities in older patients, which increases the labelling complexity for both the automatic and human labeller.

\begin{figure*}[]
    \centering
    \includegraphics[width=0.83\linewidth]{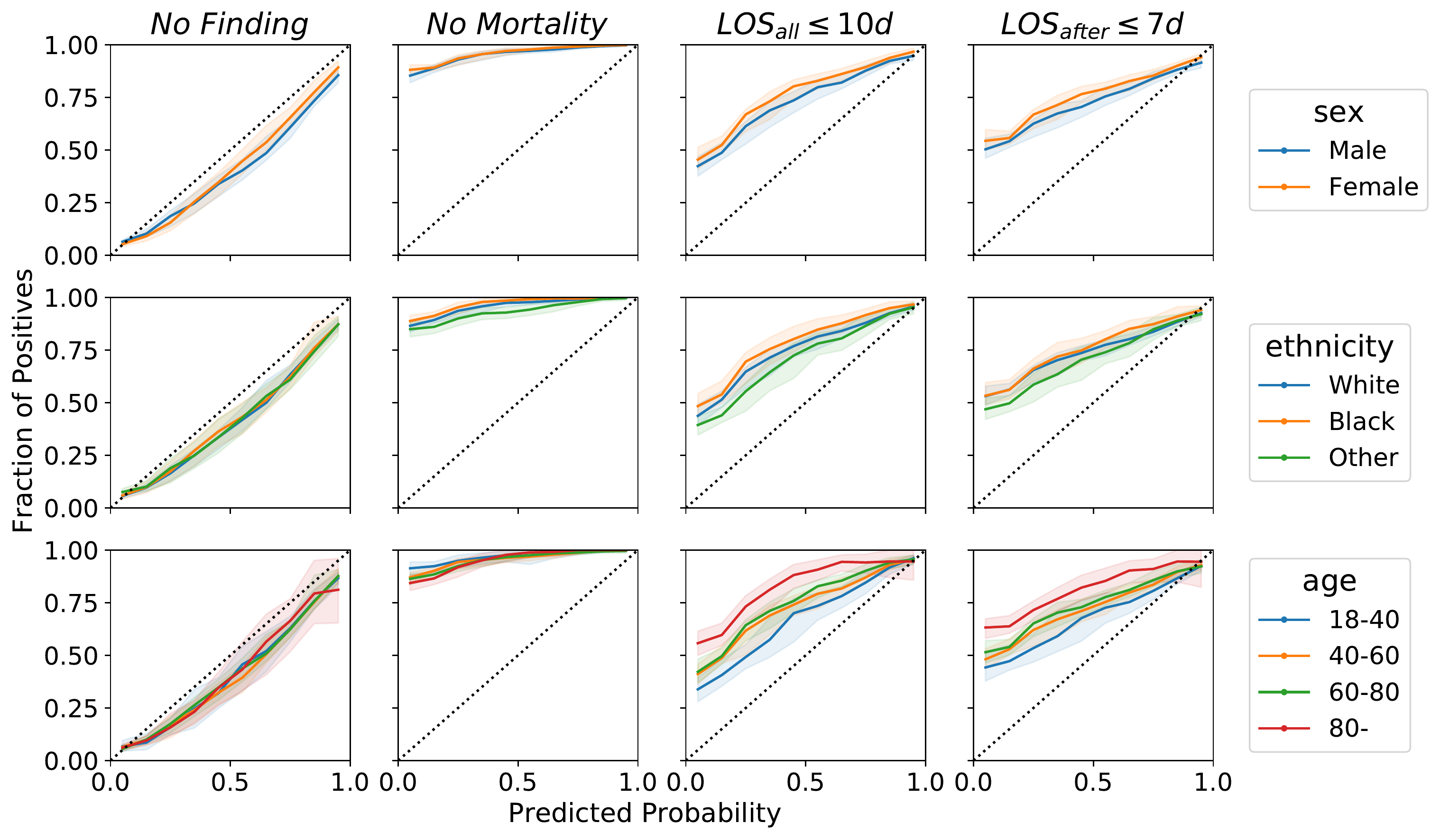} 
    \caption{Per-group calibration curves of \texttt{ERM} models trained to predict \textit{No Finding} in MIMIC-CXR, evaluated on the original \textit{No Finding} task and three proxy labels. Error bounds correspond to 95\% confidence intervals from 250 bootstrap iterations.  }
    \label{fig:proxy_labels_calibration}
\end{figure*}

\subsection{Proxy Labels}

\paragraph{Motivation} 
Due to the poor quality of the \textit{No Finding} labels, evaluation on the basis of those labels may mask consequential differences in the properties of the model across groups.
As a proof-of-concept, we conduct a case-study where we consider evaluation on the basis of \textit{proxy labels} that can be reliably measured using electronic health records (EHRs). 
The key analysis adopts the set-up of \citet{obermeyer2019dissecting} and evaluates differences across groups in the calibration of the \textit{No Finding} classifier against the proxy labels.
This analysis is motivated to assess label bias on the assumption that the distribution of the proxy label does not differ across groups conditioned on the true probability of \textit{No Finding}.

\paragraph{Setup} We link x-rays in MIMIC-CXR with metadata in MIMIC-IV, and select a subset of x-rays which were taken during each patient's hospital stay. We then construct three proxy tasks, which we expect to correlate positively and highly with \textit{No Finding}: (1) \textit{No Mortality}: the patient does not die in hospital; (2) \textit{LOS\textsubscript{all} $\leq$ 10d}: the total length of the patient's hospital stay is less than 10 days; (3) \textit{LOS\textsubscript{after} $\leq$ 7d}: the patient's length of hospital stay after the x-ray's acquisition time is less than 7 days. 
Further descriptive statistics can be found in Appendix \ref{app:proxy_labels}. 

\paragraph{Results} We show per-group calibration curves for each proxy label in Figure \ref{fig:proxy_labels_calibration}, and performance metrics in Appendix Figure \ref{fig:proxy_labels_performance}. 
Overall, performance of the models assessed against the proxy labels is greatly reduced compared to the performance against the \textit{No Finding} labels, but this is not surprising given that the models were not trained to predict the proxy labels.
The calibration curves indicate global overestimation of risk assessed against the \textit{No Finding} labels, with no significant differences in the calibration curves across groups.
While this result is replicated when assessing calibration against the proxy mortality label when stratifying by ethnicity, we observe greater mortality rates conditioned on the predicted probability of \textit{No Finding} for those with ``Other'' ethnicity, and observe some stratification by age. Evaluation on the basis of the proxy length of stay labels indicates a lower probability of a shorter length of stay conditioned on the predicted probability of \textit{No Finding} for patients with ``Other'' ethnicity. 
We further observe significant stratification in the conditional rates of prolonged length of stay across age groups, but it is unclear whether such differences are due to expected differences in care delivery on the basis of age or due to dataset biases in the \textit{No Finding} label. This proof-of-concept study demonstrates that label bias in the training data may lead to disparities in group calibration when the \textit{No Finding} classifier is deployed in clinical settings. %

\section{Discussion}
\label{sec:discussion}

\subsection{On Performance}
From Section \ref{sec:results}, we find that methods which seek to improve worst-case group performance do not outperform simple data balancing. We also find that methods which achieve equalized odds do so by worsening performance for all groups. Overall, we conclude that current computational methods are not effective at improving the fairness of chest x-ray classifiers in a useful manner over simple baselines.

\subsection{On Definitions of Fairness}
In this work, we explored the minimax definition of fairness, which has several advantages over conventional group fairness definitions. First, any minimax fair classifier can be made into a group-fair classifier by systematically worsening performance for non-worst-case groups \citep{martinezBlindParetoFairness2021}. However, turning an arbitrary group-fair classifier into a minimax fair classifier is a much more difficult task. Second, we note that there are many impossibility theorems that exist between group fairness definitions \citep{del2020review, liuImplicitFairnessCriterion2019}, and so inherent conflicts exist between the metrics which one may wish to enforce fairness with respect to (e.g. calibration metrics, PPV, TPR, and FPR). Such impossibility theorems do not exist for minimax fairness, though careful selection of the error function is still required. 

We note that debiasing a model to satisfy equalized odds typically also results in a change in the effective decision threshold applied to one or more groups, as a result of explicit threshold adjustment induced by post-processing \citep{hardt2016equality} or through the miscalibration induced by procedures such as \texttt{MMDMatch}.
As a consequence, decisions made on the basis of models debiased for equalized odds are unlikely to be made at thresholds that are concordant with clinical practice guidelines \citep{grundy20192018} or at thresholds that were selected on the basis of preferences for potential outcomes or the effectiveness of the clinical intervention associated with the model \citep{wynants2019three,corbett-daviesMeasureMismeasureFairness2018,bakalarFairnessGroundApplying2021}.

\subsection{On Sources of Bias}

In this work, we examined biases conceived of as properties of a machine learning model in Section \ref{sec:results}, as well as investigated the degree of label bias in \textit{No Finding} in Section \ref{sec:dissecting_bias}. Our results in Section \ref{sec:dissecting_bias} show that label bias is a significant issue in MIMIC-CXR, which may be partially responsible for the observed performance disparities between age groups. Such biases in automatically-generated labels may be partially ameliorated through the use of contextual language models which have improved agreement with the radiologist standard \citep{smit2020chexbert, mcdermott2020chexpert++}. However, this would only resolve the label bias that exists in extracting label information from radiology reports. Additional biases could exist in how radiologists write their reports and deduce their conclusions from the radiographs (e.g. as a result of cognitive biases) \citep{busby2018bias}.

In addition, there are many biases external to the machine learning pipeline which we do not consider \citep{suresh2019framework}. For example, it has been shown that Black patients are less likely than White patients to receive diagnostic imaging in the emergency department after adjusting for a variety of covariates \citep{ross2020influence}, an example of measurement error \citep{jacobs2021measurement}. Further details about the mechanism of these biases would be required to conduct relevant sensitivity analyses, but such details cannot be extracted solely from the observed data without further assumptions.

\subsection{Best Practices for Fairness in Clinical Settings}

\paragraph{Evaluate comprehensively.} First, we recommend evaluating per-group performance over a wide range of metrics. Examining a larger set of metrics across operating thresholds gives a holistic view of where gaps between protected groups lie. We specifically emphasize calibration error as an evaluation metric which has been deemed important in clinical risk scores \citep{crowson2016assessing, alba2017discrimination, antoniou2021evaluation}, but is relatively underexplored in the clinical \textit{fairness} literature. Differing calibration curves between protected groups means that deployment at any fixed operating threshold would result in differing implied thresholds on the true risk between protected groups \citep{foryciarz2021evaluating}. 

\paragraph{Consider sources of bias in the data.} Even after a comprehensive evaluation, the metrics obtained are only as valid as the dataset they have been evaluated on. It is crucial to consider any potential biases in the data generating mechanism and how they could shift the dataset distribution away from the real-world deployment distribution. Where possible, steps should be taken to correct such biases (e.g. collecting additional data in a fairness-aware way). %

\paragraph{Not all gaps need to be corrected.}
When observing disparities in performance, it is critical to think about the data generating process to determine whether these gaps are clinically justified. For example, the task could be inherently more difficult for some groups (e.g. older populations due to comorbidities). Blindly trying to equalize performance in these cases could lead to worse welfare for all \citep{hu2020fair}. 

We stress that computation alone is insufficient to ensure that the use of machine learning in healthcare is equitable or does not introduce harm \citep{mccradden2020ethical,chen2020ethical}. Assessing and mitigating potential harms ultimately requires reasoning about the sources of health disparities and the capacity of the intervention that the model informs to address them.

\paragraph{Which model to choose?}

Prior work has advocated for the deployment of the model that produces the most statistically accurate estimate of risk, and applying the same risk threshold regardless of group membership \citep{corbett-daviesMeasureMismeasureFairness2018}. As such a model is most likely to be an \texttt{ERM} or \texttt{Balanced-ERM} model, our results seem to agree with this conclusion. We further recommend the selection of a model where no group can achieve better performance without worsening the performance of another group (i.e. Pareto optimality \citep{martinezMinimaxParetoFairness2020}). We leave the exploration of Pareto optimality of clincial classifiers to future work, along with potential theoretical justifications for our empirical findings \citep{maity2021does}.

\clearpage
\section*{Institutional Review Board (IRB)}
This research does not require IRB approval.

\acks{Haoran Zhang is supported by a grant from the Quanta Research Institute. Karsten Roth is supported by the International Max Planck Research School for Intelligent Systems (IMPRS-IS) and acknowledges his membership in the European Laboratory for Learning and Intelligent Systems (ELLIS) PhD program. Dr. Marzyeh Ghassemi is funded in part by Microsoft Research, and a Canadian CIFAR AI Chair held at the Vector Institute. Resources used in preparing this research were provided, in part, by the Province of Ontario, the Government of Canada through CIFAR, and companies sponsoring the Vector Institute. }

\bibliography{references}

\appendix
\counterwithin{figure}{section}
\counterwithin{table}{section}

\clearpage
\onecolumn

\FloatBarrier
\section{Group Fairness Definitions}
\label{apd:group_fair}

\begin{table*}[!hbtp]
\centering
\caption{Commonly used group fairness definitions, the conditional independence statements that they entail, and the metric which they equalize in the binary classification setting. Here, $\hat{Y}, Y \in \{0, 1\}.$\label{tab:group_fair_mets}}
\begin{tabular}{@{}lll@{}}
\toprule
\textbf{Name}                            & \textbf{Independence Statement} & \textbf{Equalized Metric} \\ \midrule
Demographic Parity                       & $\hat{Y} \ind G $                                  & Predicted prevalence      \\
Equality of Odds                         &  $\hat{Y} \ind G\ \mid\ Y $                           & TPR, FPR                  \\
Equality of Opportunity (Positive Class) &       $\hat{Y} \ind G\ \mid\ Y = 1$                           & TPR                       \\
Equality of Opportunity (Negative Class) &                             $\hat{Y} \ind G\ \mid\ Y = 0$        & FPR                       \\
Predictive Parity                        &                             $Y \ind G\ \mid\ \hat{Y}=1$        & PPV                       \\ \bottomrule
\end{tabular}
\end{table*}

\noindent In Table \ref{tab:group_fair_mets}, we present several commonly used group fairness definitions assuming a binary prediction. When $S = h(\bm{x}) \in [0, 1]$ is instead a risk score, one definition that has been used is \textit{probabilistic} equalized odds \citep{pleissFairnessCalibration2017}: 
\begin{align*}
    \forall y \in \{0, 1\}: \E_{(\bm{x}, y) \sim g_1}[h(\bm{x}) \mid Y = y] = \E_{(\bm{x}, y) \sim g_2}[h(\bm{x}) \mid Y = y]
\end{align*}

Another important fairness criteria for a risk score is that it is calibrated for all groups. Mathematically, this is defined as \citep{pleissFairnessCalibration2017}: 

\begin{align*}
    \forall g \in G: \forall p \in [0, 1]: P_{(\bm{x}, y) \sim g} [y = 1 \mid h(\bm{x}) = p] = p
\end{align*}

\section{Cohort Statistics}\label{apd:cohort_stats}

\begin{table}[!h]
    \centering
    \begin{tabular}{@{}lrr@{}}
\toprule
                     & \textbf{MIMIC-CXR} & \textbf{CheXpert} \\ \midrule
\textbf{Location}    & Boston             & Stanford     \\ \midrule
\textbf{\# Images}   & 376,206             & 222,792       \\
\textbf{\# Patients} & 65,152              & 64,427        \\
\textbf{\# Frontal}  & 242,754             & 190,498       \\
\textbf{\# Lateral}  & 133,452             & 32,294        \\ \midrule
\textbf{Male}        & 52.22\%            & 59.35\%      \\
\textbf{Female}      & 47.78\%            & 40.66\%      \\ \midrule
\textbf{White}       & 60.66\%            & 56.39\%      \\
\textbf{Black}       & 15.62\%            & 5.37\%       \\
\textbf{Other}       & 23.72\%            & 38.24\%      \\ \midrule
\textbf{18-40}       & 14.75\%            & 13.88\%      \\
\textbf{40-60}       & 32.35\%            & 31.07\%      \\
\textbf{60-80}       & 39.41\%            & 39.01\%      \\
\textbf{80-}         & 13.49\%            & 16.05\%      \\ \bottomrule
\end{tabular}
    \caption{Summary statistics for MIMIC-CXR and CheXpert. Percentages shown correspond to the fraction of the population belonging to a particular group. Note that we use all images in our experiments.}
    \label{tab:cohort_stats_1}
\end{table}

\begin{table}[!h]
    \centering
\begin{tabular}{@{}lrrrrrr@{}}
\toprule
                 & \multicolumn{3}{c}{\textbf{MIMIC-CXR}}                          & \multicolumn{3}{c}{\textbf{CheXpert}}                           \\ \cmidrule(lr){2-4} \cmidrule(lr){5-7}
                 & \textbf{No Finding} & \textbf{Fracture} & \textbf{Pneumothorax} & \textbf{No Finding} & \textbf{Fracture} & \textbf{Pneumothorax} \\ \midrule
\textbf{Male}    & 37.09\%             & 1.88\%            & 4.00\%                & 9.88\%              & 4.48\%            & 8.98\%                \\
\textbf{Female}  & 42.62\%             & 1.46\%            & 2.77\%                & 10.22\%             & 3.42\%            & 8.25\%                \\ \midrule
\textbf{White}   & 34.60\%             & 1.98\%            & 4.04\%                & 9.40\%              & 4.45\%            & 9.11\%                \\
\textbf{Black}   & 44.29\%             & 0.74\%            & 1.81\%                & 11.68\%             & 2.44\%            & 5.75\%                \\
\textbf{Other}   & 49.87\%             & 1.54\%            & 2.85\%                & 10.70\%             & 3.68\%            & 8.46\%                \\ \midrule
\textbf{18-40}   & 63.41\%             & 1.02\%            & 3.58\%                & 20.49\%             & 4.26\%            & 12.48\%               \\
\textbf{40-60}   & 45.51\%             & 1.65\%            & 3.20\%                & 12.40\%             & 3.98\%            & 8.63\%                \\
\textbf{60-80}   & 31.91\%             & 1.75\%            & 3.68\%                & 7.00\%              & 3.60\%            & 8.91\%                \\
\textbf{80-}     & 22.86\%             & 2.25\%            & 2.93\%                & 3.70\%              & 5.09\%            & 4.95\%                \\ \midrule
\textbf{Overall} & 39.73\%             & 1.68\%            & 3.41\%                & 10.02\%             & 4.05\%            & 8.68\%                \\ \bottomrule
\end{tabular}
    \caption{Prevalence of each label for each group in MIMIC-CXR and CheXpert.}
    \label{tab:cohort_stats_2}
\end{table}

\FloatBarrier

\section{Additional Training Details}\label{apd:train_det}
\paragraph{Optimization} We train all models using the Adam optimizer using a batch size of 64 and a learning rate of $10^{-4}$ on NVIDIA Tesla P100 GPUs. We evaluate on the validation fold every 200 steps, and stop training if the worst-case ROC has not improved for 5 such evaluations. 

\paragraph{Image Augmentations} We apply the following image augmentations to the training set: random flipping of the images along the horizontal axis, random rotation of up to 10 degrees, and a crop of a random size (75\% - 100\%) and a random aspect ratio (3/4 to 4/3). 

\paragraph{Hyperparameters} We search over the following hyperparameter space for each of the methods:
\begin{itemize}
    \item \texttt{Adversarial}: $\alpha \in \{0.01, 0.05, 0.1, 1.0, 2.0, 5.0, 20.0, 30.0, 50.0, 100.0\}$
    \item \texttt{MMDMatch}: $\lambda \in \{0.1, 0.5, 0.75, 1.0, 2.0, 3.0, 4.0, 5.0, 10.0, 20.0, 30.0, 50.0, 100.0\}$
     \item \texttt{MeanMatch}: $\lambda \in \{0.1, 0.5, 0.75, 1.0, 2.0, 3.0, 4.0, 5.0, 10.0, 20.0, 30.0, 50.0, 100.0\}$
       \item \texttt{GroupDRO}: $\eta \in \{0.01, 0.1, 1.0\}$
     \item \texttt{FairALM}: $\eta \in \{10^{-1}, 10^{-2}, 10^{-3}\}$
     \item \texttt{JTT}: $\lambda_{up} \in \{2, 3, 5, 10, 30, 50\}$
\end{itemize}

\FloatBarrier

\clearpage
\section{Additional Experimental Results - Main Experimental Grid}\label{apd:results}
\subsection{Pneumothorax Prediction in MIMIC-CXR}

\begin{figure*}[!h]
    \centering
    \hspace*{-1.cm}\includegraphics[width=1.15\textwidth]{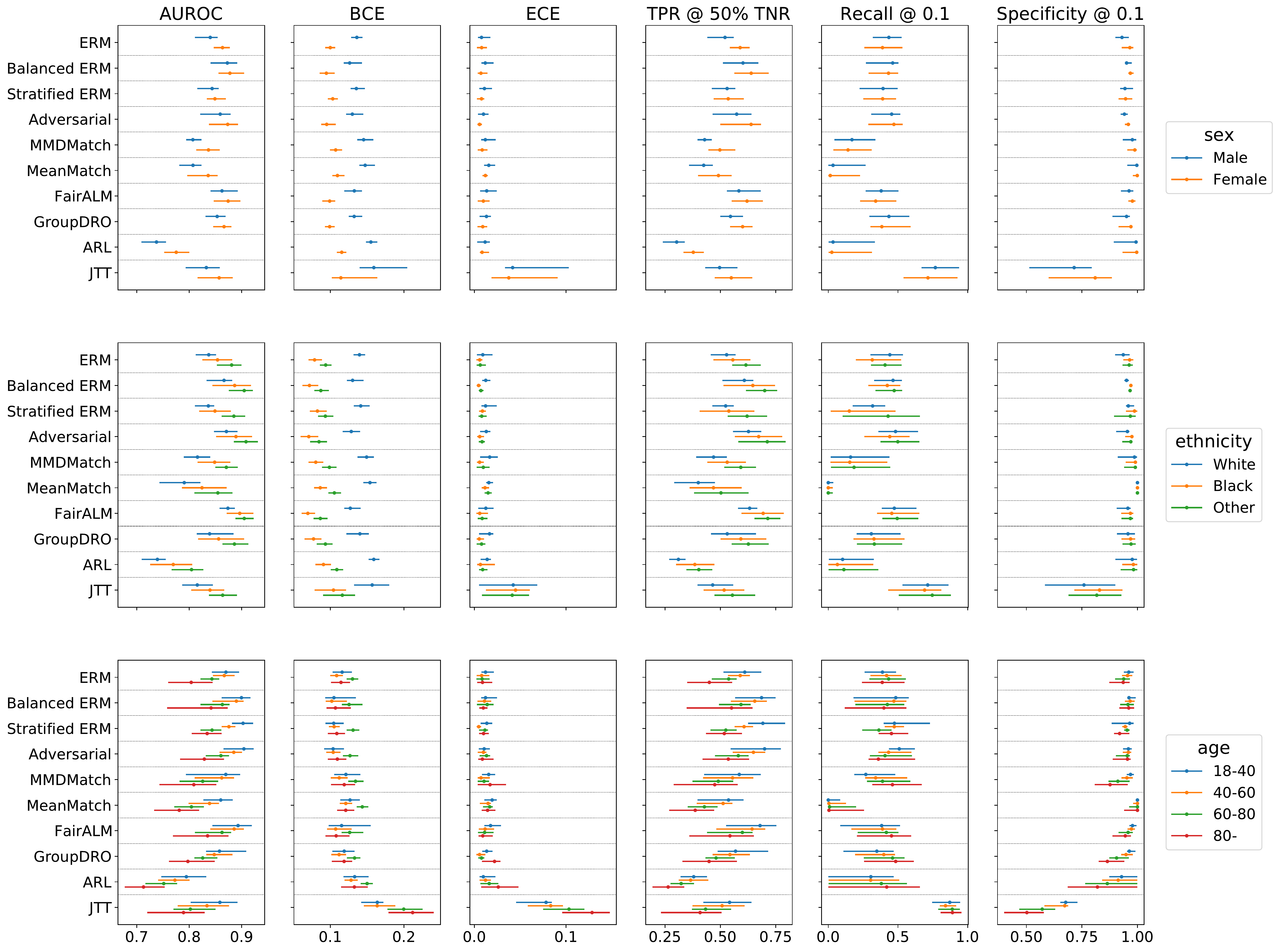} 
    \caption{Comparison of the performance of models that predict \textit{Pneumothorax} in MIMIC-CXR. Error bars indicate 95\% confidence intervals from 250 bootstrap iterations.}
\end{figure*}

\begin{figure*}[!h]
    \centering
    \hspace*{-1cm}\includegraphics[width=1.15\textwidth]{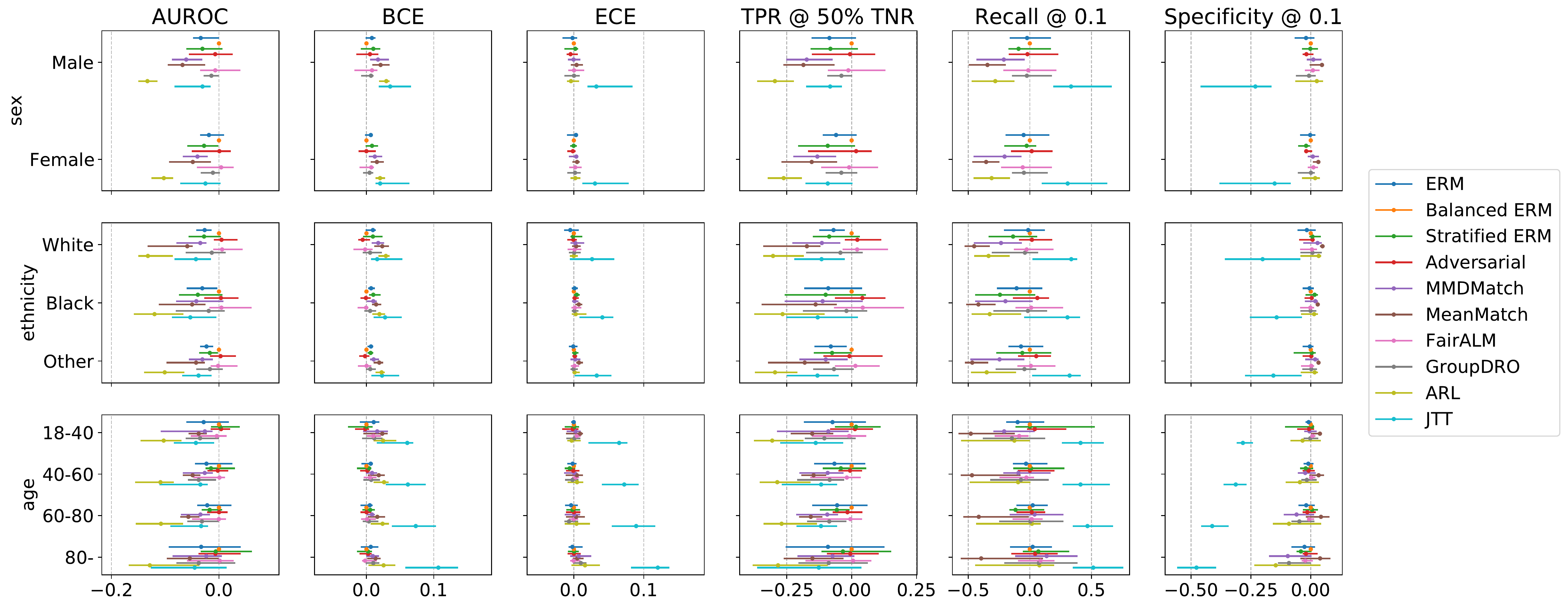} 
    \caption{Comparison of models that predict \textit{Pneumothorax} in MIMIC-CXR. We show the difference in performance between each model and \texttt{Balanced ERM}. Error bars indicate 95\% confidence intervals from 250 bootstrap iterations.}
\end{figure*}

\FloatBarrier \clearpage
\subsection{Fracture Prediction in MIMIC-CXR}

\begin{figure*}[!h]
    \centering
    \hspace*{-1.cm}\includegraphics[width=1.15\textwidth]{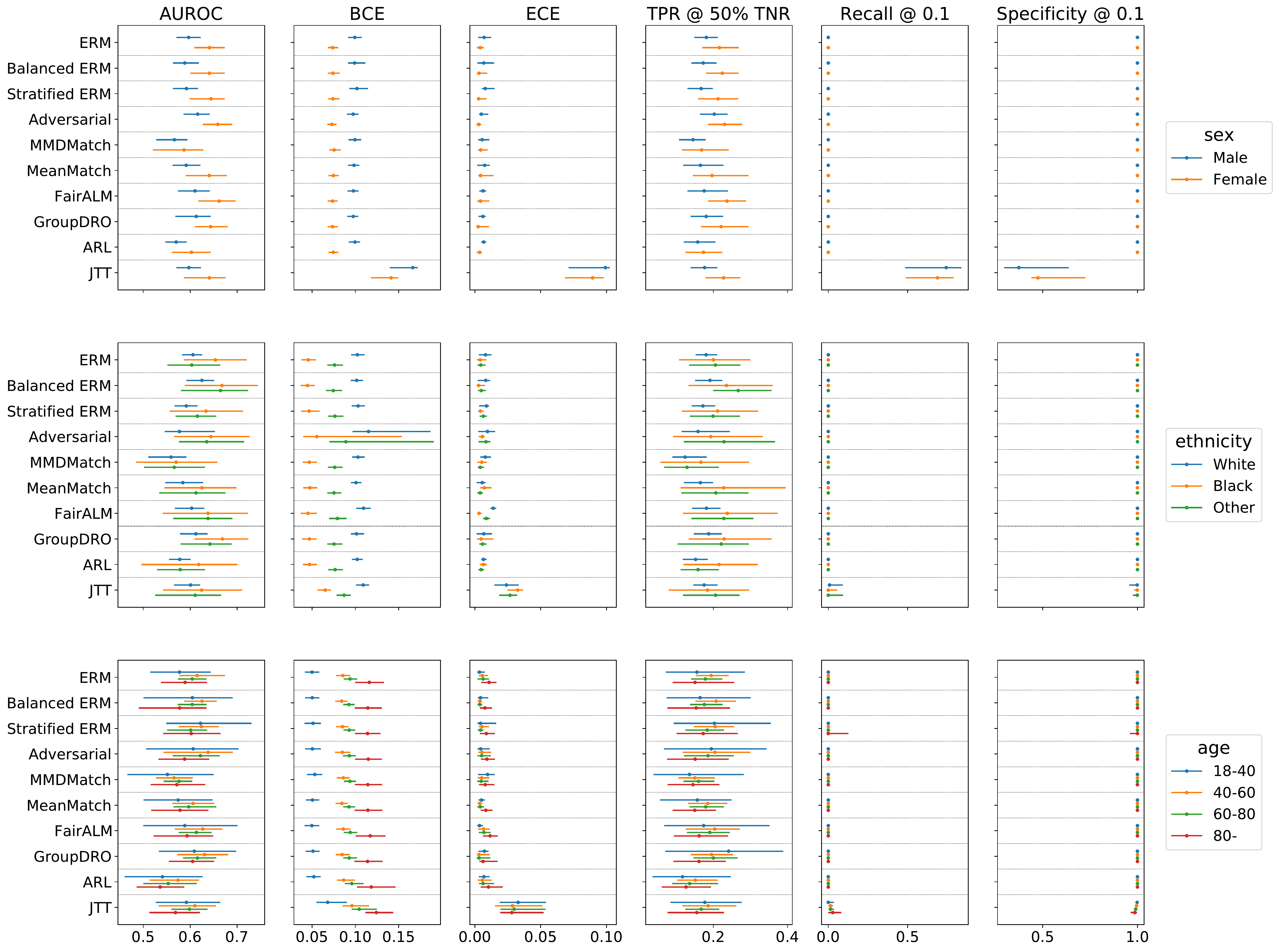} 
    \caption{Comparison of the performance of models that predict \textit{Fracture} in MIMIC-CXR. Error bars indicate 95\% confidence intervals from 250 bootstrap iterations.}
\end{figure*}

\begin{figure*}[!h]
    \centering
    \hspace*{-1cm}\includegraphics[width=1.15\textwidth]{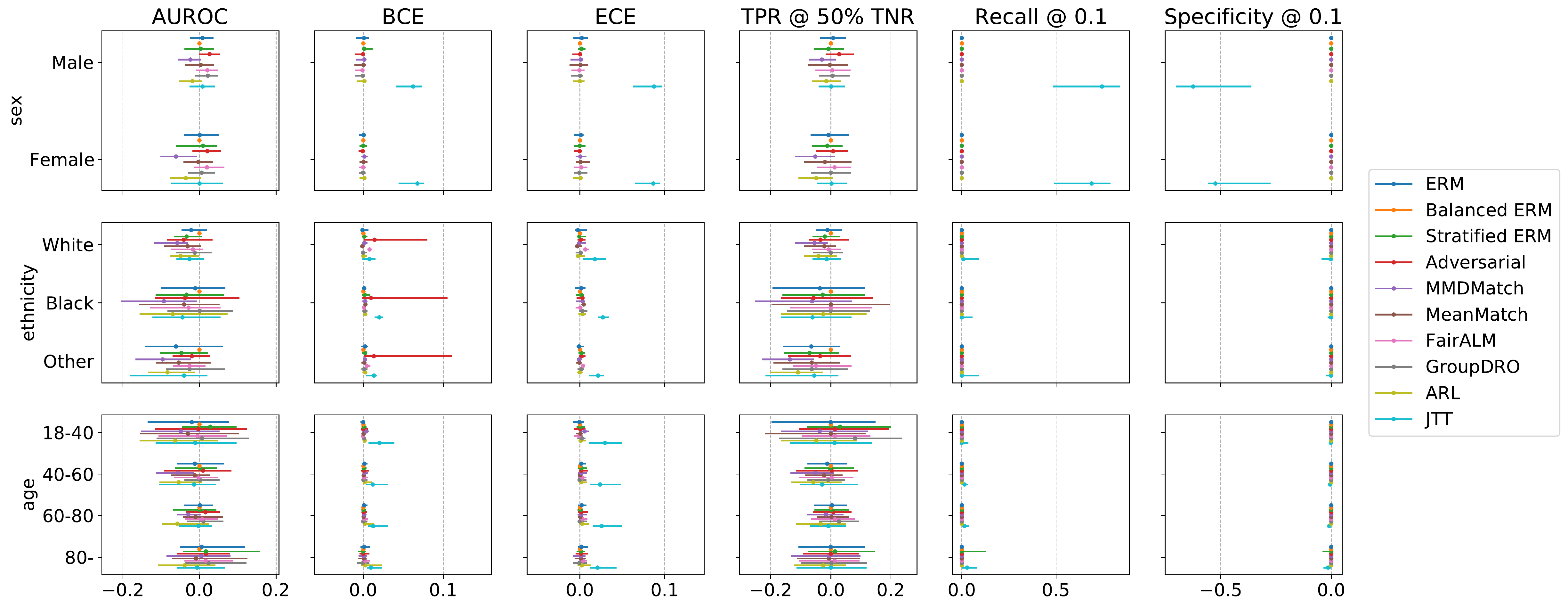} 
    \caption{Comparison of models that predict \textit{Fracture} in MIMIC-CXR. We show the difference in performance between each model and \texttt{Balanced ERM}. Error bars indicate 95\% confidence intervals from 250 bootstrap iterations.}
\end{figure*}

\FloatBarrier \clearpage

\subsection{No Finding Prediction in CheXpert}

\begin{figure*}[!h]
    \centering
    \hspace*{-1cm}\includegraphics[width=1.15\textwidth]{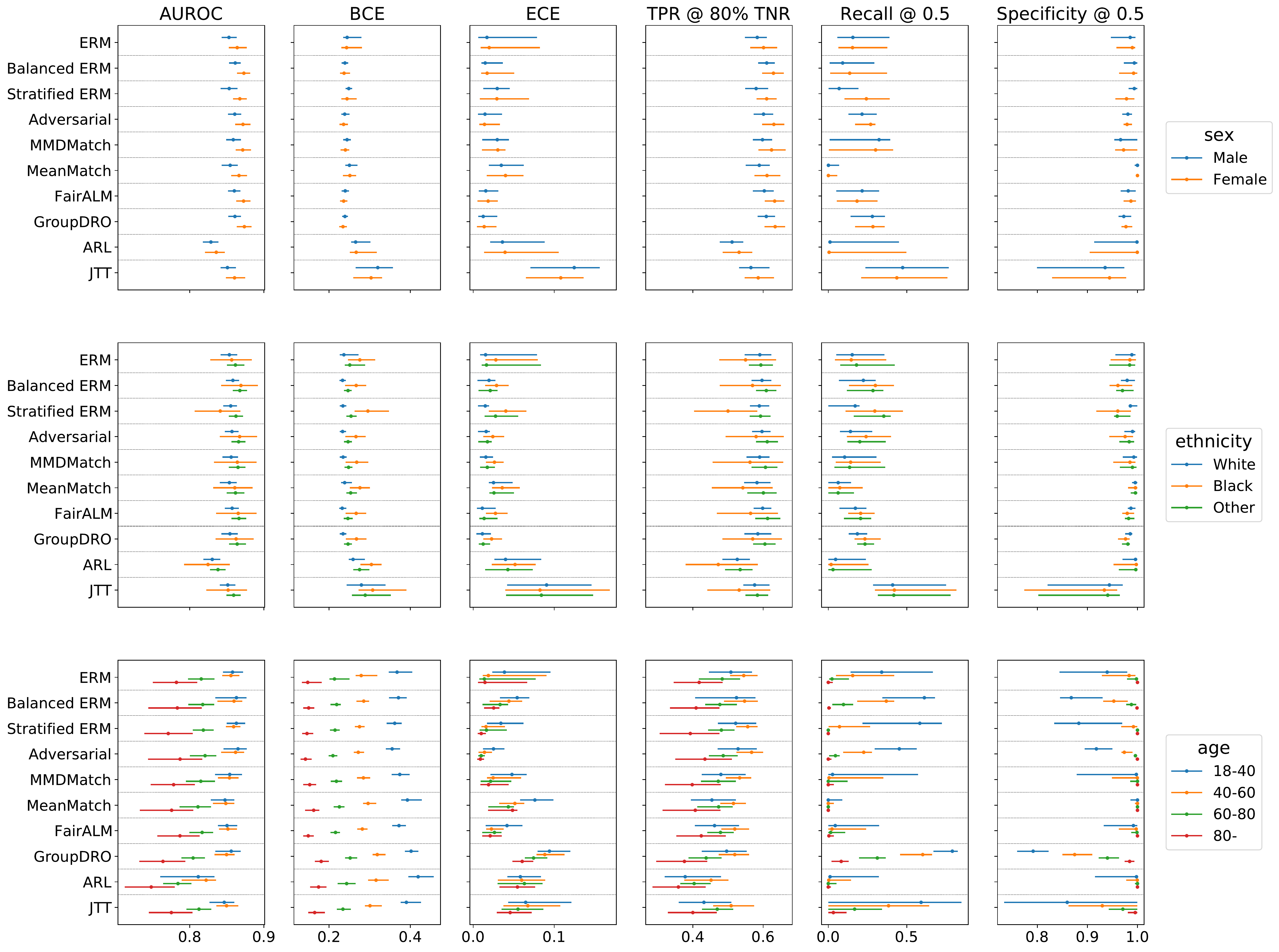} 
    \caption{Comparison of the performance of models that predict \textit{No Finding} in CheXpert. Error bars indicate 95\% confidence intervals from 250 bootstrap iterations.}
\end{figure*}

\begin{figure*}[!h]
    \centering
    \hspace*{-1cm}\includegraphics[width=1.15\textwidth]{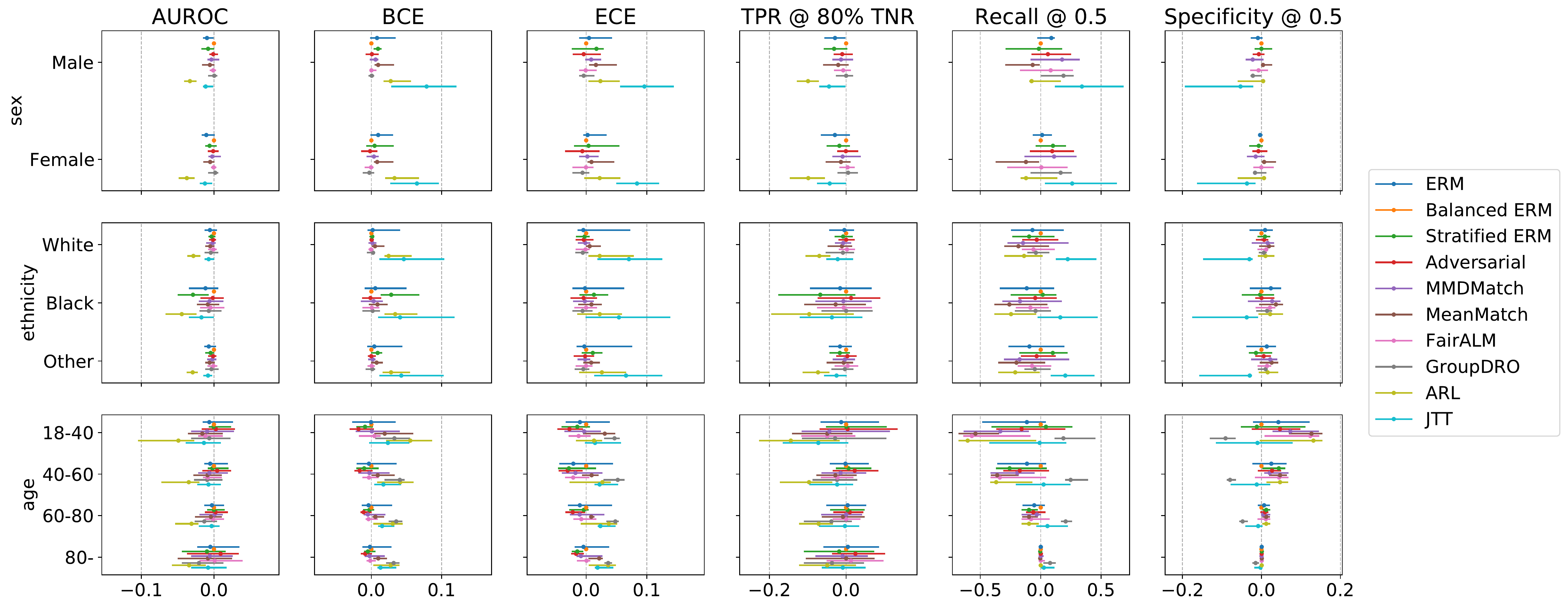} 
    \caption{Comparison of models that predict \textit{No Finding} in CheXpert. We show the difference in performance between each model and \texttt{Balanced ERM}. Error bars indicate 95\% confidence intervals from 250 bootstrap iterations.}
\end{figure*}

\FloatBarrier \clearpage
\subsection{Pneumothorax Prediction in CheXpert}

\begin{figure*}[!h]
    \centering
    \hspace*{-1cm}\includegraphics[width=1.15\textwidth]{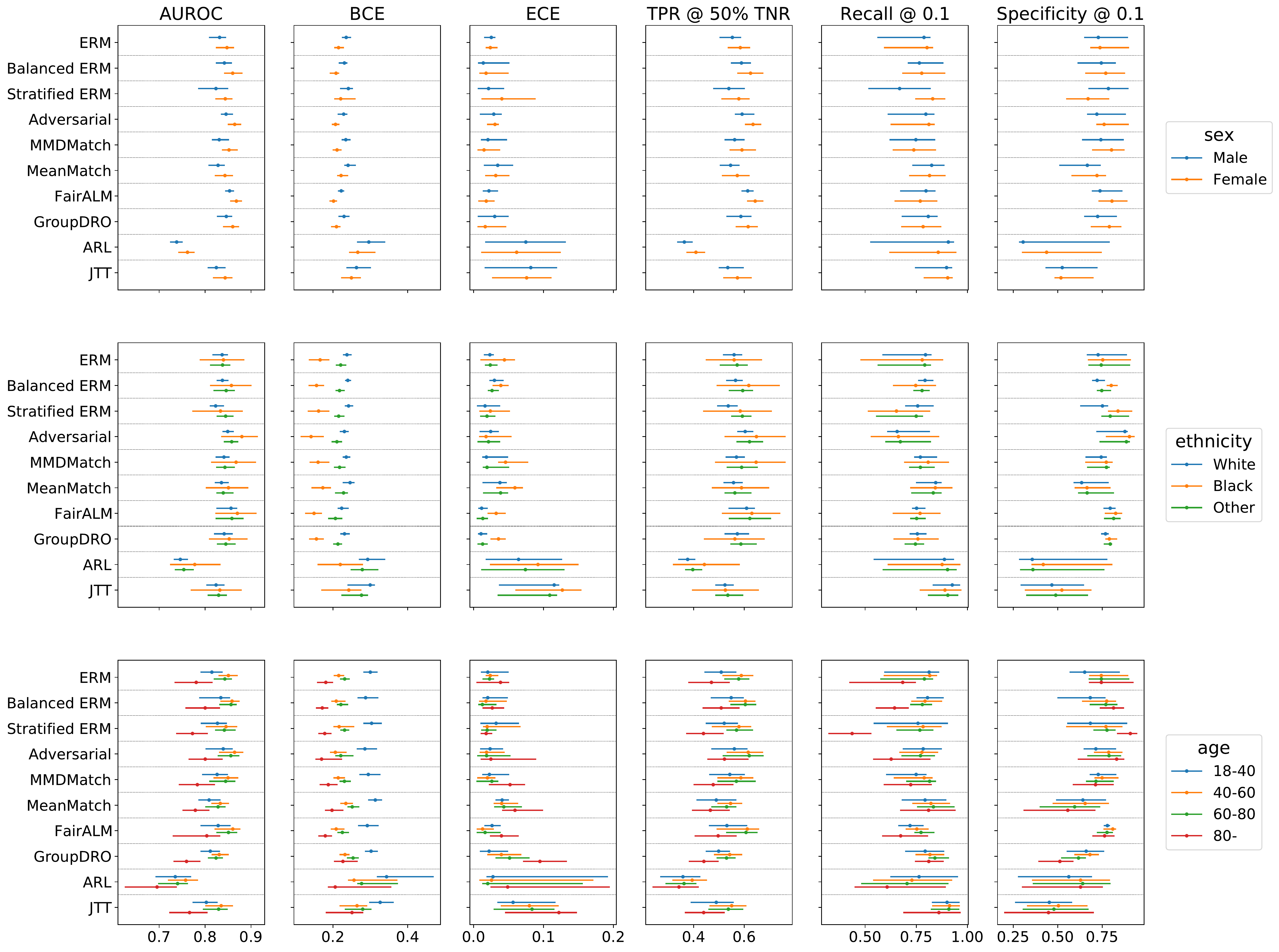} 
    \caption{Comparison of the performance of models that predict \textit{Pneumothorax} in CheXpert. Error bars indicate 95\% confidence intervals from 250 bootstrap iterations.}
\end{figure*}

\begin{figure*}[!h]
    \centering
    \hspace*{-1cm}\includegraphics[width=1.15\textwidth]{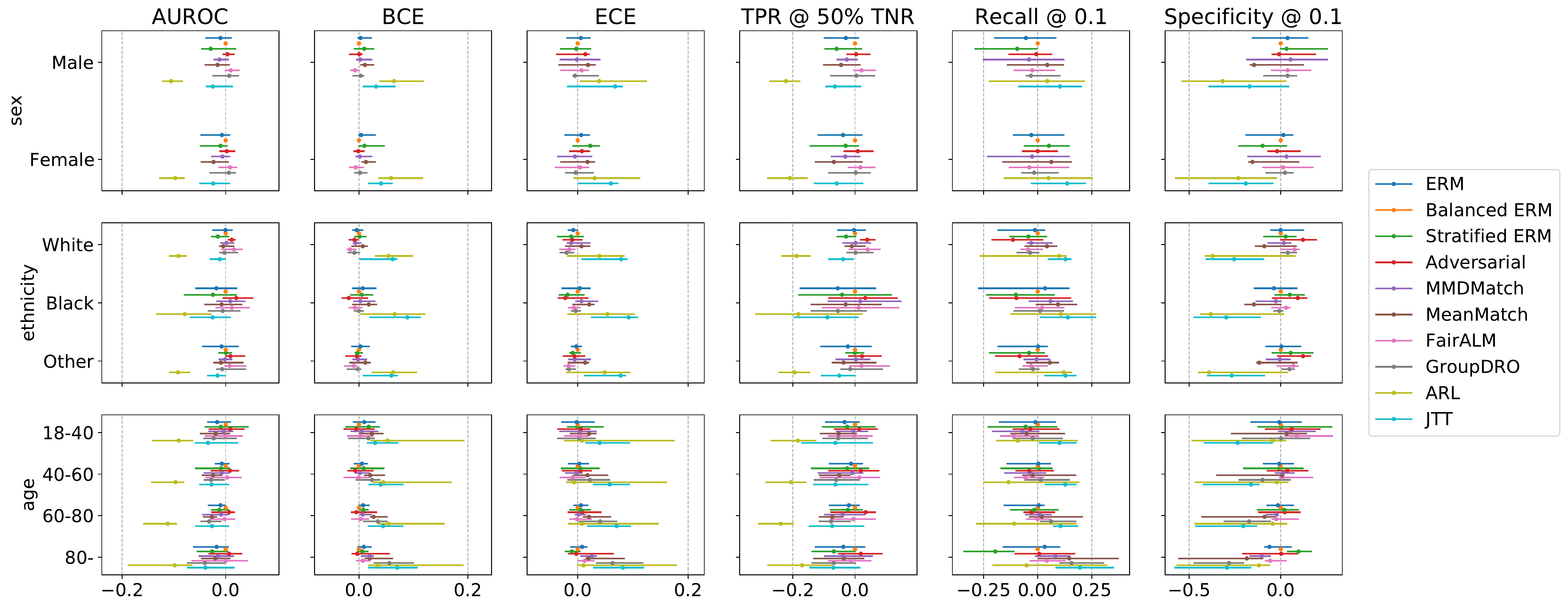} 
    \caption{Comparison of models that predict \textit{Pneumothorax} in CheXpert. We show the difference in performance between each model and \texttt{Balanced ERM}. Error bars indicate 95\% confidence intervals from 250 bootstrap iterations.}
\end{figure*}

\FloatBarrier \clearpage
\subsection{Fracture Prediction in CheXpert}

\begin{figure*}[!h]
    \centering
    \hspace*{-1cm}\includegraphics[width=1.15\textwidth]{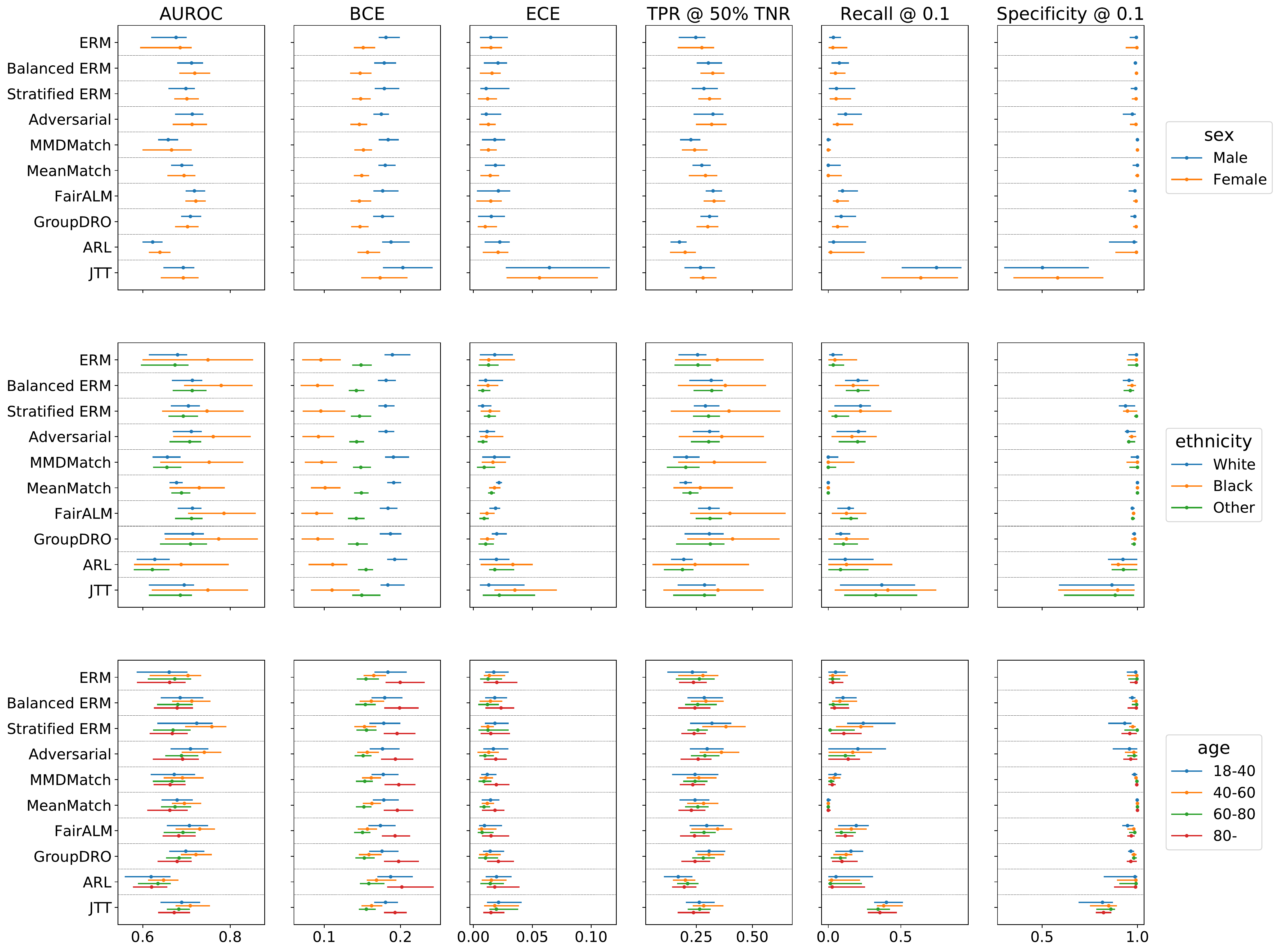} 
    \caption{Comparison of the performance of models that predict \textit{Fracture} in CheXpert. Error bars indicate 95\% confidence intervals from 250 bootstrap iterations.}
\end{figure*}

\begin{figure*}[!h]
    \centering
    \hspace*{-1cm}\includegraphics[width=1.15\textwidth]{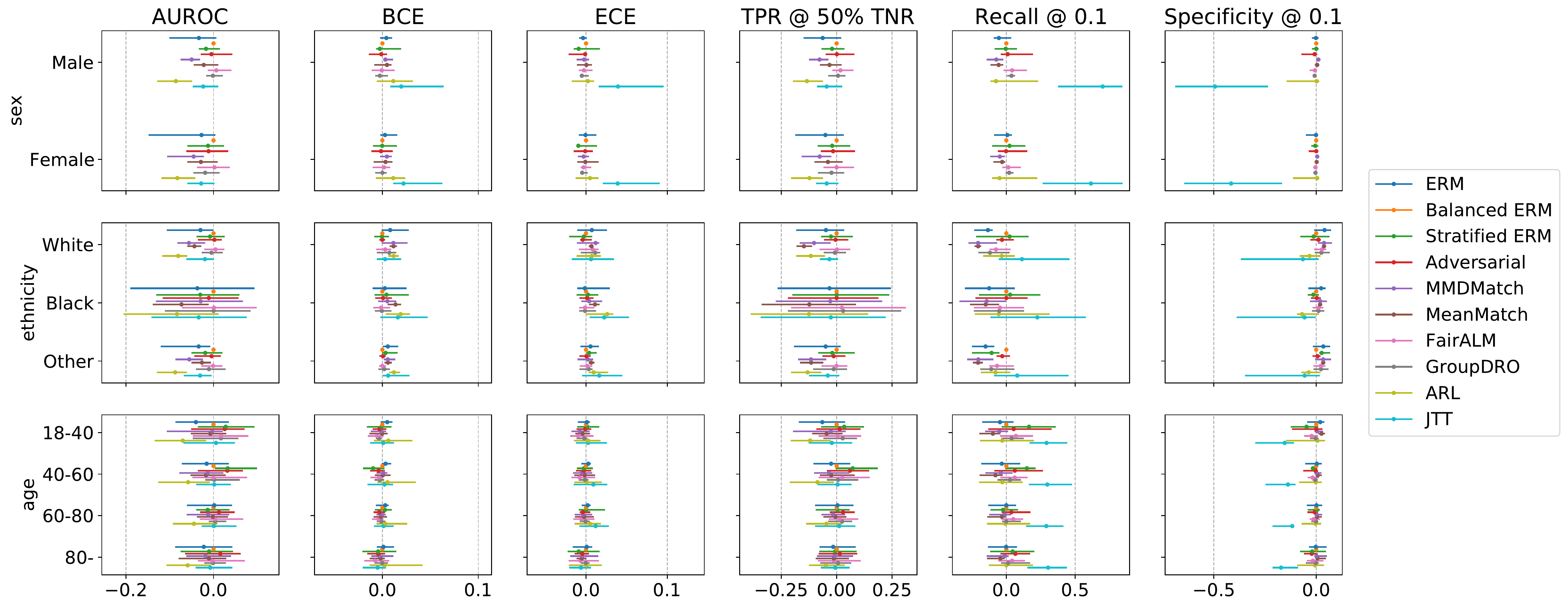} 
    \caption{Comparison of models that predict \textit{Fracture} in CheXpert. We show the difference in performance between each model and \texttt{Balanced ERM}. Error bars indicate 95\% confidence intervals from 250 bootstrap iterations.}
\end{figure*}

\FloatBarrier

\clearpage
\section{Additional Experimental Results - Label Bias Dissection}

\subsection{Radiologist-Labelled Samples}

\begin{figure*}[!h]
    \centering
    \hspace*{-1.cm} \includegraphics[width=1.15\linewidth]{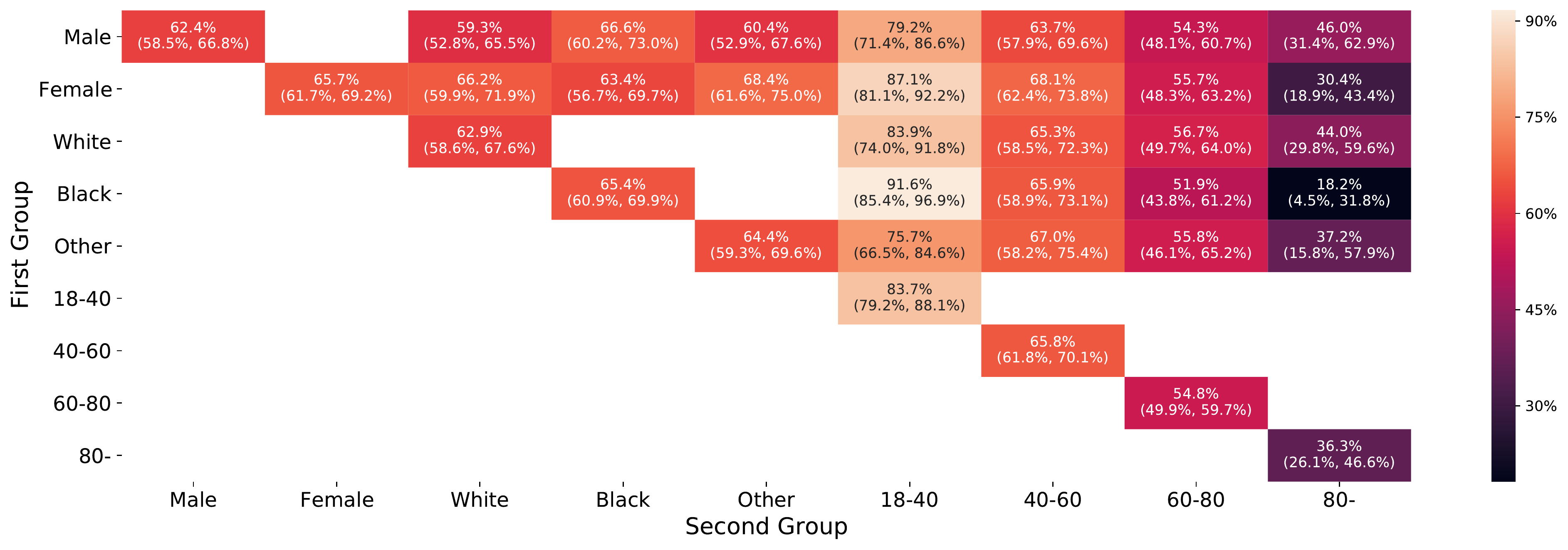} 
    \caption{Accuracy of the CheXpert labeller on 1,200 radiology reports which it labels as \textit{No Finding} relative to the radiologist gold standard, for each protected group and intersectional subgroup. Error bounds shown are 95\% confidence intervals obtained through 500 bootstrap iterations. }
    \label{fig:rad_labels_with_CIs}
\end{figure*}

\subsection{Proxy Labels}
\label{app:proxy_labels}
\begin{table}[!h]
    \centering
    \begin{tabular}{@{}rrrrrr@{}}
\toprule
\multicolumn{1}{l}{}                 & \multicolumn{1}{c}{\textbf{Proportion}} & \multicolumn{4}{c}{\textbf{Prevalence}}                                                                                    \\ \cmidrule(lr){2-2}  \cmidrule(lr){3-6} 
\multicolumn{1}{l}{}                 & \textbf{}                               & \textbf{No Finding} & \textbf{No Mortality} & \textbf{LOS\textsubscript{all} $\leq$ 10d} & \textbf{LOS\textsubscript{after} $\leq$ 7d} \\ \midrule
\textbf{Male}                        & 55.00\%                                 & 28.84\%             & 92.28\%                       & 63.93\%                          & 65.76\%                           \\
\textbf{Female}                      & 45.00\%                                 & 31.54\%             & 93.56\%                       & 69.55\%                          & 70.36\%                           \\ \midrule
\textbf{White}                       & 66.56\%                                 & 28.71\%             & 92.88\%                       & 66.22\%                          & 67.98\%                           \\
\textbf{Black}                       & 15.88\%                                 & 35.54\%             & 95.50\%                       & 73.37\%                          & 72.12\%                           \\
\textbf{Other}                       & 17.56\%                                 & 30.22\%             & 90.38\%                       & 61.10\%                          & 63.37\%                           \\ \midrule
\textbf{18-40}                       & 10.57\%                                  & 45.50\%             & 96.56\%                       & 69.25\%                          & 69.82\%                           \\
\textbf{40-60}                       & 31.81\%                                 & 36.18\%             & 94.35\%                       & 68.31\%                          & 68.82\%                           \\
\textbf{60-80}                       & 42.41\%                                 & 25.78\%             & 91.93\%                       & 62.74\%                          & 64.89\%                           \\
\textbf{80-}                         & 15.22\%                                 & 18.48\%             & 89.75\%                       & 71.01\%                          & 72.53\%                           \\ \midrule
\multicolumn{1}{l}{\textbf{Overall}} &                                         & 30.06\%             & 92.86\%                       & 66.46\%                          & 67.83\%                           \\ \bottomrule
\end{tabular}
    \caption{Summary statistics for the modified cohort containing only x-rays from MIMIC-CXR that were taken during a patient's hospital stay. We show the proportion of each protected group in the data, as well as their prevalence for the original \textit{No Finding} label and three proxy labels derived from MIMIC-IV. The total size of the dataset is 42,877 images. }
    \label{tab:proxy_label_stats}
\end{table}

\begin{figure*}[!h]
    \centering
    \hspace*{-1.cm} \includegraphics[width=1.15\linewidth]{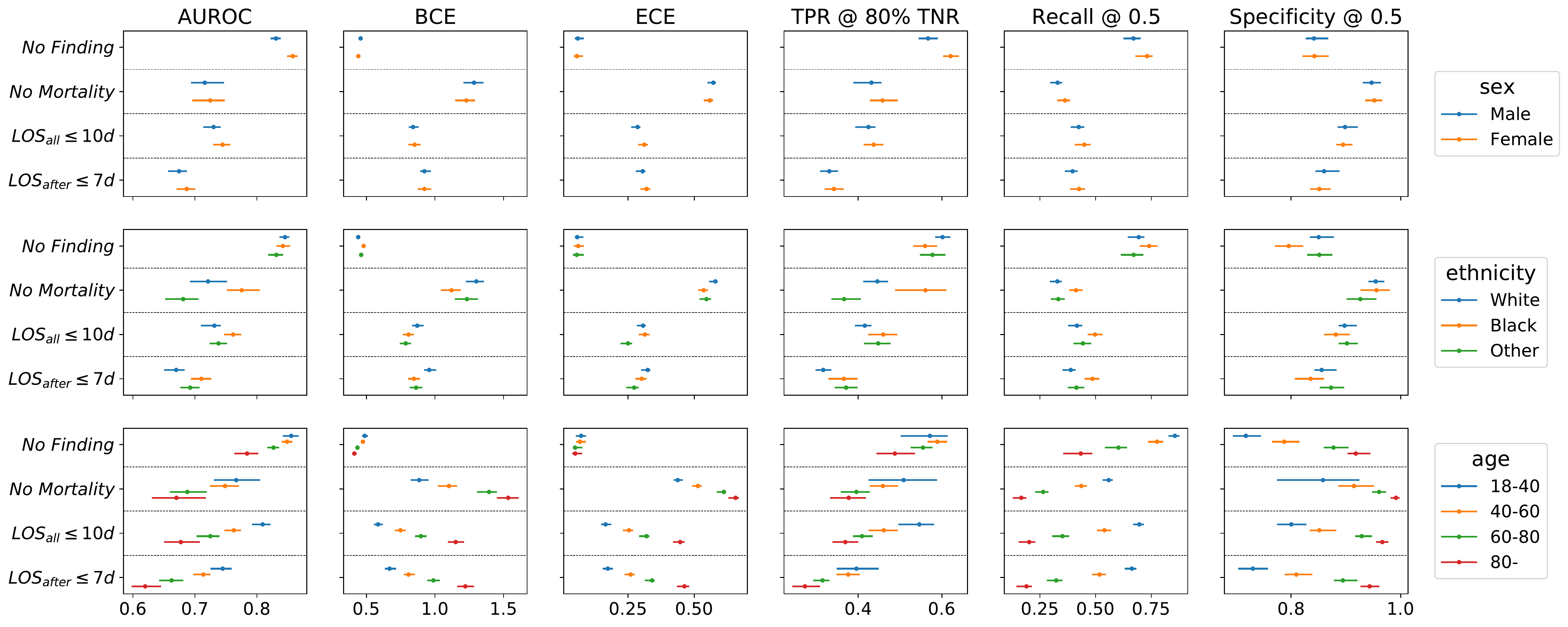} 
    \caption{Performance of an \texttt{ERM} model trained to predict \textit{No Finding} in MIMIC-CXR, evaluated on the original \textit{No Finding} task and three proxy labels. Error bounds correspond to 95\% confidence intervals from 250 bootstrap iterations.  }
    \label{fig:proxy_labels_performance}
\end{figure*}

\end{document}